\documentclass[lettersize,journal]{IEEEtran}
\usepackage{amsmath,amsfonts}
\usepackage{amssymb}
\usepackage{txfonts}
\usepackage{multirow}
\usepackage{algorithmic}
\usepackage{array}
\usepackage[caption=false,font=normalsize,labelfont=sf,textfont=sf]{subfig}
\usepackage{textcomp}
\usepackage{stfloats}
\usepackage{url}
\usepackage{verbatim}
\usepackage{graphicx}
\def\BibTeX{{\rm B\kern-.05em{\sc i\kern-.025em b}\kern-.08em
    T\kern-.1667em\lower.7ex\hbox{E}\kern-.125emX}}
\usepackage{balance}
\begin{document}
\title{Domain Generalized Recaptured Screen Image Identification Using SWIN Transformer}
\author{Preeti Mehta, Aman Sagar, Suchi Kumari
}

\markboth{Journal of \LaTeX\ Class Files,~Vol.~18, No.~9, September~2020}%
{How to Use the IEEEtran \LaTeX \ Templates}

\maketitle

\begin{abstract}
	
An increasing number of classification approaches have been developed to address the issue of image rebroadcast and recapturing, a standard attack strategy in insurance frauds, face spoofing, and video piracy. However, most of them neglected scale variations and domain generalization scenarios, performing poorly in instances involving domain shifts, typically made worse by inter-domain and cross-domain scale variances. 
To overcome these issues, we propose a cascaded data augmentation and SWIN transformer domain generalization framework (DAST-DG) in the current research work
Initially, we examine the disparity in dataset representation. A feature generator is trained to make authentic images from various domains indistinguishable. This process is then applied to recaptured images, creating a dual adversarial learning setup. Extensive experiments demonstrate that our approach is practical and surpasses state-of-the-art methods across different databases. Our model achieves an accuracy of approximately 82\% with a precision of 95\% on high-variance datasets.
	
\end{abstract}


\section{Introduction}
\label{introduction}
\begin{figure}[!htbp]
	\centering
	\includegraphics[width=8cm,height=4cm]{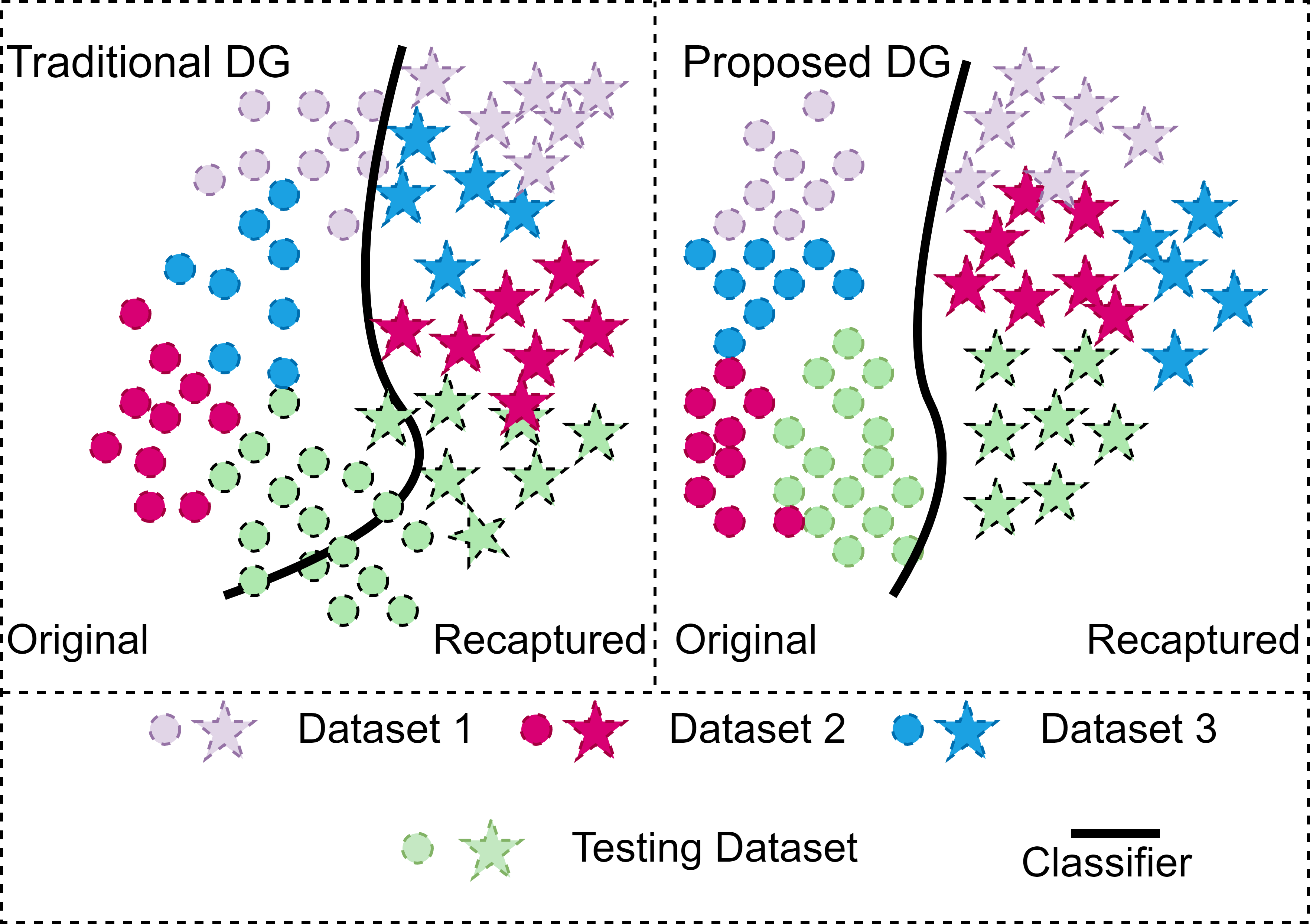}
	\caption{Left: Traditional methods of dataset domain generalization place source domains with acquiring a common feature space. Still, they cannot obtain a selective class boundary on the testing dataset. Right: Our DAST-DG method custer all the original image samples while separating the recaptured image sets from various domains to learn a class boundary.}
	\label{fig1}
\end{figure}

With technological advancement, digital multimedia files can nowadays be simply recorded by cameras and shared over the Web. Tampering, including reacquiring, is an immediate threat to digital image integrity. As a result, for images to act as reliable witnesses, their originality must be thoroughly verified \cite{mehta2022near,anjum2020recapture}. According to \cite{cao2010identification,mahdian2015detecting}, humans have difficulty differentiating between the two classes—i.e., recaptured and original images. Recaptured images can deceive real-world systems, particularly forensic systems, leading to Rebroadcast Image Attacks (RIA). To mitigate such fraud and attacks, it is essential to incorporate detection forensics to identify rebroadcast images. In RIA detection, handcrafted features include wavelet statistical distributions \cite{zhai2013recaptured,kim2016face,zhang2018face}, noise analysis, and color and texture non-uniformity \cite{gao2010single}. Among all the available RIA detection techniques, texture distribution is considered a reliable solution \cite{mehta2022CNN}. Physical artefacts such as specularity, blurriness, and chromaticity are effective differential features.
In addition to handcrafted features, some neural network methods \cite{edmunds2018face,thongkamwitoon2015image,ke2013image,jung2015recaptured} are explored to improve classifier performance. Besides, various databases are generated for model training to build a robust classification model, which extracted artefacts from the input images having different domains because of the variance in features, including scale, illumination, and color \cite{torralba2011unbiased}. Therefore, the domain generalization of the datasets is a practical challenge in RIA tasks. Till now, most of the proposed frameworks achieve good performance for single-domain scenarios. Thus, as illustrated in Fig. \ref{fig1}, a cross-domain recaptured image detection task is presented to gain an understanding of a shared feature space for classifying intra, inter and cross-domain scale variances. Domain Generalization (DG) techniques offer direct solutions for this problem.

\subsection{Motivation}
\label{motivation}

The existing recaptured image detection techniques mainly fall under two main categories: machine learning and deep learning. Both approaches identify characteristic features without the requirement for additional information embedding. 
For case in point, a questioned digital image can go through scrutinizing for inherent artefacts for tampering detection. However, the existing techniques lack robustness against all available datasets. After the acquisition procedure, the LCD recaptured image goes through an outright image acquisition chain with or without minimal or post-processing (tampering). The existing tampering detection techniques consider this image as an original copy. However, most existing recapture detection schemes focus on a single dataset. This paper focuses on the domain generalization problem for digital image datasets recaptured through camera-screen communication. 

\subsection{Major Contributions}
\label{contribution}
Before proposing the authentication technique for recaptured detection, we want to emphasize some significant points and the key contributions of the proposed work:
\begin{itemize}
	\item First, the features of dataset images are extracted by the SWIN transformer. The loss between the original and predicted labels is calculated using the traditional binary cross entropy. 
	\item The proposed model integrates the global and local feature maps using the SWIN transformer architecture.
	\item To evaluate the effectiveness of our proposed approach, we tested our model for intra, inter, and cross-database examination using a combination of three databases. 
	\item We consider various experiment settings, including deep learning models, loss functions, and training and testing datasets. 
	\item Experimental results show that the proposed scheme with SWIN transformer with softmax activation function and cross-entropy loss outperformed other approaches. Specifically, under the most challenging scenario of cross-domain experiment, i.e., evaluation across different types of images produced by different devices and environmental settings, the proposed neural network has achieved an accuracy rate of around 86\%.
\end{itemize}

\subsection{Paper Organization}
\label{organization}
After a comprehensive review of related works in Section \ref{related work}, Section \ref{preliminary} delineates the the definition of domain generalization and SWIN transformer. Section \ref{methodology} provides the detail of the model used in our research, providing an in-depth description of the experiments conducted with distinct models, elaborating on their architecture and configurations. In Section \ref{result}, we rigorously evaluate the models using specific performance metrics, presenting a detailed analysis of their results. Finally, Section \ref{conclusion} concludes the paper by summarizing the key findings and outlining potential avenues for future research in this domain.

\section{Related Work}
\label{related work}

This section provides a concise overview of recent advancements in recaptured image detection and related work. Numerous existing methods rely on prior knowledge of artefacts in reacquired images, utilizing this information to classify images based on these identified artefacts. researchers proposed numerous methods for near-duplicate image detection considering a variety of artefact types.
Table \ref{tab:summary} summarizes those methods for near-duplicate image detection based on extracted artefacts, providing insights into the techniques employed by different authors in this field.

\begin{table*}[!htbp]
	\centering
	\caption{Summarization of Methods for Near-Duplicate Image Detection based on Artefacts}
	\label{tab:summary}
	\resizebox{15cm}{!}{%
		\begin{tabular}{|p{2.5cm}| p{2.5cm}| p{3.5cm}| p{3.5cm}| p{2cm}|}
			\hline
			\textbf{Artefact Type} & \textbf{Methods} & \textbf{Key Findings/Results} & \textbf{Limitations/Challenges} & \textbf{References} \\ 
			\hline
			Aliasing  \ref{alias}              & - Multi-LBP, Multi-scale Wavelet-statistical features & - Extraction of aliasing, loss of detail, and color distortion\newline - Achieved accuracies up to 98.95\% & - Difficulty in removal post-processing\newline - Limited effectiveness if used alone & \cite{cao2010identification,patel2015live,muammar2013investigation,mahdian2015detecting} \\ \hline
			Blurriness    \ref{blurriness}          & - Block and blurriness effects from JPEG compression\newline - Wavelet decomposition & - Extraction of blur characteristics\newline - Achieved accuracies up to 98.58\% & - Limited applicability to JPEG format images & \cite{thongkamwitoon2015image,li2015effective,luan2017face,anjum2020recapture,mehta2022near} \\ \hline
			Noise     \ref{noise}              & - Features based on noise and double JPEG compression\newline - Periodic patterns from LCD monitors & - Detection of noise characteristics\newline - Techniques for noise reduction and classification & - Not rotationally invariant & \cite{yin2012digital,ke2013image,jung2015recaptured,wang2017simple} \\ \hline
			Contrast/Color/Texture Non-Uniformity \ref{contrast}& - Tone response function adjustments\newline - Texture feature extraction using methods like Local Binary Pattern (LBP) descriptor & - Correction of contrast and color balance errors\newline - Texture feature extraction for classification & - Reliability of color moment extraction & \cite{ke2013image,moreira2013exploring,kim2016face,edmunds2018face,kose2012classification,luan2017face,yu2008recaptured,bai2010physics,gao2010single,zhai2013recaptured,kim2016face,zhang2018face} \\ \hline
			Deep Learning      \ref{AFE}     & - CNN models with pre-processing layers\newline - Utilization of learned filters instead of pre-defined kernels & - Automatic feature extraction from original and recaptured images\newline - Improved detection accuracy & - Computational complexity & \cite{mittal2021deep,yang2016recapture,li2017image,choi2017content,yue2020recaptured,zhu2022.1recaptured,luo2021scale,mehta2022wavelet} \\ 
			\hline
		\end{tabular}%
	}
\end{table*}
The detailed description of the artefacts (in Table \ref{tab:summary}) is provided in the following subsections.
\subsection{Aliasing Artefacts}
\label{alias}

The aliasing artefacts, commonly referred to as colour moir\'{e} \cite{niu2021morie}, pose challenges in post-processing but can be mitigated with proper setup during recapturing \cite{thongkamwitoon2015image}. 
The frequency response of recaptured images, typically contains more high-frequency components than original images, primarily due to additional noise. As a result, aliasing and loss-of-detail effects are more pronounced in recaptured images.

\subsection{Blurriness Artefacts}
\label{blurriness}

An authentic image captures a diverse range of colours and edges with varying levels of contrast and sharpness. However, when the same scene is recaptured with a digital device, inherent blurring inevitably occurs. This blurriness persists despite correct focusing by the acquisition device, primarily due to imperfections in the capturing lens, including spherical aberration leading to the barrel and pincushion distortion \cite{visentini2013modelling}. Additionally, blur distortion can be introduced by the camera acquisition pipeline, resulting in unique blur characteristics for each camera \cite{thongkamwitoon2015image}.

\subsection{Noise Artefacts}
\label{noise}

The noise distribution in recaptured media is influenced by factors such as the characteristics of the recaptured camera, the brightness setting of the imitating medium, and surrounding conditions. While the noise characteristic of the camera can accurately identify the source camera, its limitation lies in its lack of rotational invariance.

\subsection{Contrast, Colour and Texture Non-Uniformity Artefacts}
\label{contrast}

Modern digital cameras, LCD screens, and projector devices adhere to the sRGB colour encoding standard. Digital cameras typically apply a tone response function during pre-processing, deviating slightly from the standard sRGB response to produce visually pleasing images with slightly higher contrast. However, in screen-camera communication processes, the resulting media includes the tone response functions of both the camera and the screen, leading to higher overall contrast than single captures.
Colour-related artefacts in recaptured media may include errors in colour balance, such as tints introduced during recapture from projectors or LCD screens and increased colour saturation. Authors often incorporate colour moments and contrast features in their training methodologies \cite{ke2013image,moreira2013exploring,kim2016face,edmunds2018face,kose2012classification}. However, the reliable extraction of colour moments depends on the analysed content. Additionally, some authors have utilized specularity distribution as another feature for colour classification \cite{luan2017face,yu2008recaptured,bai2010physics,gao2010single}.
Physical-based features for recapture detection, as explored in \cite{gao2010single}, revealed that colour histogram and contrast were the most effective features, while specularity and blurriness were less convincing. Texture features have proven to be effective characteristics for recaptured media \cite{yin2012markov,zhai2013recaptured,kim2016face,zhang2018face}, with the LBP descriptor and its variants being particularly promising for texture information extraction.

\subsection{Automatic Extracted Artefacts}
\label{AFE}

The increasing popularity of deep learning networks and their excellent detection accuracy in computer vision enable researchers to explore this technique for recognition and classification problems \cite{mittal2021deep}. 

Table \ref{table 2.2} provides a clear comparison and analysis of the features and model used for classification work of the existing recaptured image detection approaches. Additionally, the shortcomings of state-of-the-art methods are discussed.

\begin{table*}[!htbp]
	\caption{Summarization of different image recaptured detection techniques}
	\label{table 2.2}
	\centering
	
	\resizebox{17cm}{!}{
		\scalebox{0.9}{	
			\begin{tabular}{|c |p{5cm}| p{5cm}| p{5cm}|}
				\hline
				\textbf{Year (citation)} & \textbf{Artefacts (Features)} & \textbf{Classifier Model} & \textbf{Shortcomings}\\
					\hline
				2008 \cite{yu2008recaptured}		& Probability distribution,
				dithering patterns & Physical modeling of a
				recapturing process & The model performance is not good in
				planar surfaces in generic scenes. \\	\hline
				2010 \cite{gao2010single} & Colour distortions & Statistics-based features,
				chromaticity covariance
				matrix & Degradation process directly by the
				application of mathematical models.\\	\hline
				2010 \cite{cao2010identification} & Aliasing-like distortions’ colour
				distortion, blurriness & Blurriness, texture, noise, and
				colour features & The network performance is not good
				enough.	\\ 	\hline
				2013 \cite{ke2013image} & Texture, colour noise, difference
				histogram & Histogram of image local
				difference, colour moments & Feature extraction becomes more and more
				cumbersome.\\ 	\hline
				2013 \cite{muammar2013investigation} &  Noise residual features & Used 2DFT of noise residual
				and theory of cyclostationarity & This method works only for aliasing-free
				image datasets.\\ 	\hline
				2013 \cite{zhai2013recaptured} & Texture features & Based on texture features & only single feature is considered.\\	\hline
				2015 \cite{mahdian2015detecting} & aliasing distortion & Used 2DFT of noise residual
				and theory of cyclostationarity & Images with dark content create detection
				challenge. \\ 	\hline
				2015 \cite{thongkamwitoon2015image} & aliasing and blurriness & Dictionary approximation
				error of edge profiles & Performance is degrading with Kodak
				camera images. \\	\hline
				2015 \cite{Ni2015RecapturedIF} &color distortions & Features of colour moments
				and DCT coefficients & This method is only effective for JPEG
				images that are compressed images.\\ 	\hline
				2016 \cite{yang2016recapture} & Extract features from dataset
				using a CNN & Used CNN algorithm & color information not used.\\ 	\hline
				2016 \cite{SAAH2016RecognitionOR} & Texture, HSV colour, and
				blurriness & Used SVM classifier & Only used images taken by back-facing
				camera.\\	\hline
				2017 \cite{yang2017recaptured} & Quality-aware feature and
				histogram feature & Compression artefacts have
				been used & This method only works for JPEG
				compressed images.\\	\hline
				%				2017 \cite{Jianli2017image} & Co-occurrence matrices
				%				extracted from the residual image
				%				are used for detection & Support vector machine and
				%				ensemble classifier & Proposed network performance is not very
				%				high.\\
				2017 \cite{choi2017content} & Extract features from dataset
				using a CNN & CNN-based approach & Proposed network use only fewer
				convolutional layers.\\	\hline
				2017 \cite{li2017image} & Aliasing distortion and noise
				artefacts & Convolution and recurrent
				neural network-based & Large fully connected layers increase the
				model complexity.\\	\hline
				2018 \cite{sun2018recaptured} & Wavelet characteristics and noise
				characteristics & Mean value, variance, and
				skewness, extract noise image
				using LBP & The images have not been tested on
				different image formats.\\	\hline
				2018 \cite{zhu2018recaptured} & Structure distortions &Enhanced residual-based
				correlation coefficients & Only work for JPEG compressed images.\\	\hline
				2019 \cite{zhu2019recaptured} & LBP-coded maps & LBP features & Fixed filter layer at the input may ignore
				some useful information.\\	\hline
				2020 \cite{anjum2020recapture} & 
				images edge profile  & An image’s edge profile can be
				used to determine its structural
				level & Only consider edge profile and proposed
				method cannot identify images obtain from
				different media.\\	\hline
				2020 \cite{yue2020recaptured} & Aliasing distortion & AMNet with additive and multiplicative modules
				& The dataset consist of only LCD and mobile screenshots images.\\	\hline
				2021 \cite{zhou2020near} & Extracting local and global features using CNN & CFMs saliency maps & lack of transfer learning and supervised classification model.\\	\hline
				2021 \cite{yan2021cross} & Texture and Reflectance Characteristics &  convolutional neural network & only for certificate document.\\	\hline
				2021 \cite{zhao2021deep} &textual and background information &TENet &tamper localization of the forgery
				of texts is remained to be investigated.\\	\hline
				2021 \cite{luo2021scale} & scale alignment
				domain generalization framework  (SADG)& CNN & scale variances and domain shift problems addressed.\\	\hline
				2021 \cite{mehta2021detection} & Statistical texture features & SVM classifier & benchmark results not good\\	\hline
				2022 \cite{zhu2022recaptured} & normalized LTC (local ternary
				count) histograms of residual maps& SVM classifier & design more optimal residual maps and  the luminance and color artefacts study.\\	\hline
				2022 \cite{chen2022distortion} &spatial and spectral distortion models& CNN model& only for printing and scanning process and prior knowledge of the printer
				model is required.\\	\hline
				2022 \cite{zhu2022.1recaptured}&(LTC) of high order prediction error& SVM classifier&prediction error maps calculation strategies not optimal.\\	\hline
				2022 \cite{miao2022learning}&Edge, wavelet and artefact transform&Feature
				Disentanglement and Dynamic Fusion (FDDF) model& tested on Large-scale Real-scene Universal Recapture
				(RUR) dataset proposed by them.\\	\hline
				2022 \cite{chen2022domain}&Siamese network&network with ResNeXt101 backbone&results good for only few-shot fine-tuning document types.\\	\hline
				2022 \cite{zhu2022exposing}&correlations between color channels&constrained convolutional neural 
				network& attention 
				mechanism module missing.\\	\hline
				2022 \cite{liu2022recaptured}&intensity and 
				gradient information&generalized central difference convolution (GCDC) network& improvement scope in fusion strategy and attention module.\\	\hline
				2022 \cite{mehta2022near}&Edge Histogram Descriptors& SVM Classifier& Tested only on LCD recaptured images.\\	\hline
				2022 \cite{mehta2022wavelet} &Wavelet transform (DDD-DWT), CNN &SVM Classifier& Tested only on LCD recaptured images. \\	\hline
				2023 \cite{li2023recaptured} & combines local-feature and global-feature extraction modules&convolutional neural network (CNN) and vision transformer (ViT)& does not work well for the detailed images, such
				as small-scale image features.\\	\hline
				2023 \cite{li2023two} &frequency filter bank and multi-scale cross-attention fusion module&two-branch deep neural network&Limited data generalization and computationally complex.\\

					\hline
				
	\end{tabular}}}
\end{table*}

\section{preliminary} 
\label{preliminary}

In this section we will explain the definition of domain and domain generalization followed by the SWIN Transformer.

	\textbf{Domain:} Let $\varmathbb{X}$ and $\varmathbb{Y}$ denotes a nonempty input and output space, respectively. A domain is compose of data that are sampled from a distribution, $\varmathbb{D} = \{(x_i,y_i)^N_{i=1} \sim \mathcal{P_{XY}}\}$ where $x \in \varmathbb{X} \subset \mathbb{R}^d, y \in \varmathbb{Y} \subset \mathbb{R}$ denoting the features and label, respectively. $\mathcal{P_{XY}}$ denotes the input sample features and output label joint distribution with $\mathcal{X}$ and $\mathcal{Y}$ as random variables.
	
\vspace{2em}

	\textbf{Domain Generalization (DG):} We are given $\varmathbb{S}$ source domains (or training datasets), $\varmathbb{S}_{train} = \{\varmathbb{D}^i \mid i = 1:K\}$ where $\varmathbb{D}^i = \{(x^i_j,y^i_j)\}^{n_i}_{j=1}$ denoting the $i^{th}$ domain. The joint distributions between each pair of datasets are different, i.e., $\mathcal{P^\mathnormal{i}_{XY}} \neq \mathcal{P^\mathnormal{j}_{XY}}, 1 \leq i \neq j \leq K$. The objective of domain generalization is to learn a robust and generalized predictive model $\textit{M}: \varmathbb{X} \to \varmathbb{Y}$ from the $K$ training datasets to achieve a minimum prediction error on an unknown test dataset $\varmathbb{S}_{test}$ as given by the Eq. \ref{eq1}:
	\begin{equation}
		\label{eq1}
		\underset{h}\min \hspace{1em} \mathbb{E}_{(x,y)\in\varmathbb{S}_{test}} \hspace{1em}[l(h(x),y)], 
	\end{equation}
	where $\mathbb{E}$ is the error and $l(.,.)$ is the loss function and $\mathcal{P^\mathnormal{test}_{XY}} \neq \mathcal{P^\mathnormal{i}_{XY}}$. In our research, we have evaluated loss using Binary Cross Entropy (BCE) function. The loss function is defined in Eq. \ref{eq2}.
	
\begin{equation}
	\label{eq2}
	l_{BCE} = -\sum_{j \in C0} (1-t_j)\log(1 - \hat{p}_j)- \sum_{j \in C1} t_j \log(\hat{p}_j)
\end{equation}

%\begin{equation}
%	\label{eq3}
%	l_{FL} = -\alpha \left[\sum_{j \in C0} (1-t_j)^\beta\log(1 - \hat{p}_j)- \sum_{j \in C1} t_j^\beta \log(\hat{p}_j)\right]
%\end{equation}

where $\hat{p}_j$ is the estimated probability value and $t_j$ is the true value of original image class ($C1$) and recaptured image class ($C0$), respectively with $j$ numbers of sample images in respective classes.
% $\beta$ is the focusing parameter and $\alpha$ is a hyperparameter that set the balance between precision and recall values.

Some researchers have designed the DG techniques for face-antispoofing problems, such as MADDG \cite{shao2019multi} and SSDG \cite{jia2020single}. However, these techniques are highly customized and thus not applicable to the LCD recaptured images detection problem. The process of image reacquisition can involve various cameras and imitating LCD sources with different specifications, resulting in features that may cluster in feature space but across different domains.

\begin{figure}
	\centering
	
	\includegraphics[width=\columnwidth]{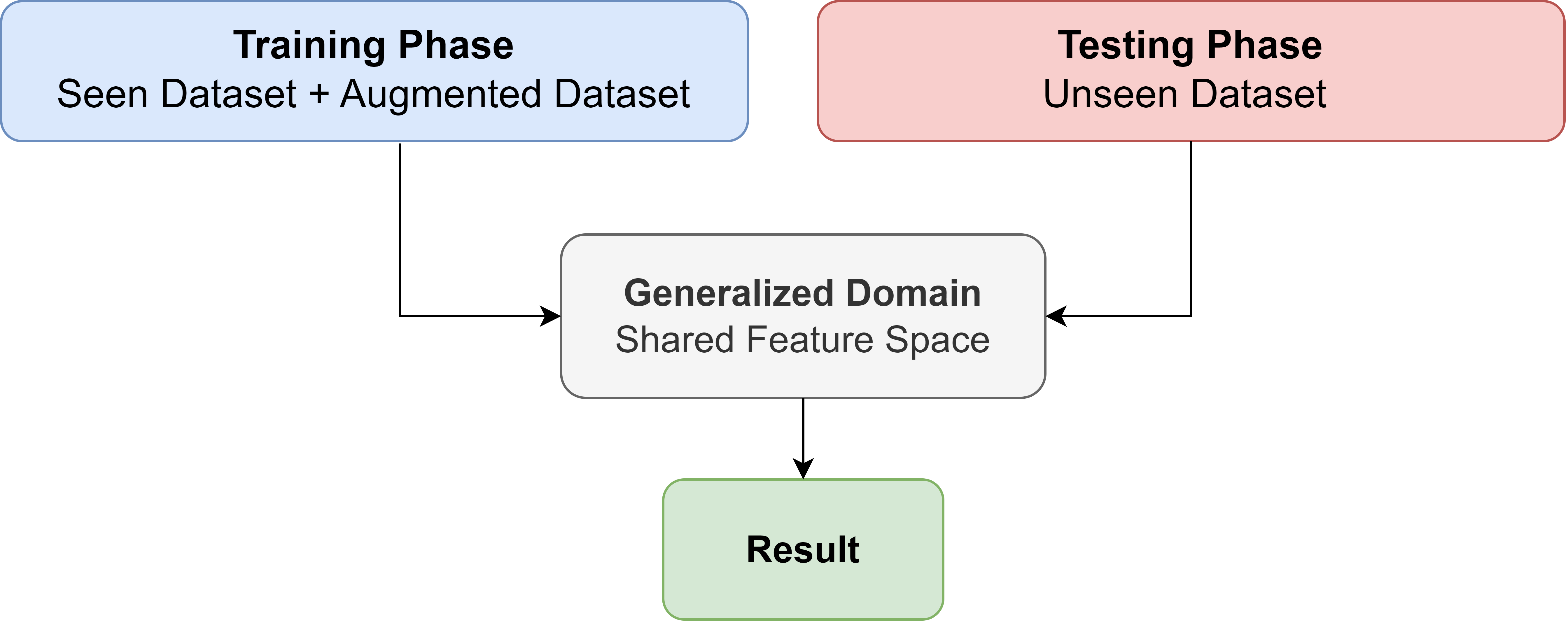}
	\caption{Introduction to inter, intra and cross-domain recapture detection. Our model aims to learn a shared feature space which is invariant to domain and scale variance setting. Both the training and testing phases contain original and recaptured images.}
	\label{fig2}
\end{figure}

Keeping the aforementioned in mind, we propose a data augmention with SWIN transformer based domain generalization framework (DAST-DG), as illustrated in Fig. \ref{fig2}. Specifically, a new testing dataset is designed from data augmentation techniques along with complex CutMix and CutOut techniques to generate more challenging datasets. The feature extraction and classification model is based on the concept of SWIN transformer. A feature generator is trained to compete with a domain discriminator to make the characteristic artefacts of original images from different domains indifferent. This will ensure that the recaptured images from other domains are separated and the original photos of all the domains are aggregated. As a result, different feature attributes can be grouped, leading to better-generalized class labelling.

\section{Proposed Methodology}
\label{methodology}
\begin{figure*}[t]
	\centering
	
	\includegraphics[width=15cm]{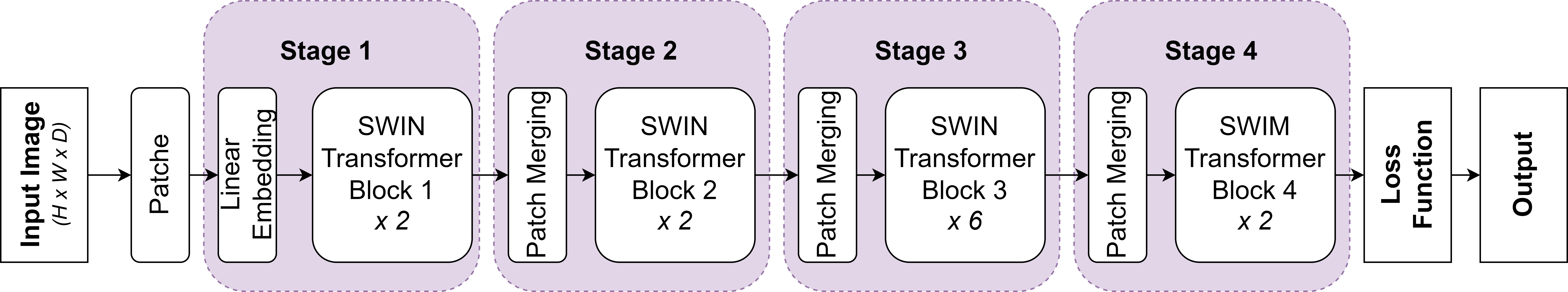}
	\caption{Architecture of the proposed SWIN transformer}
	\label{fig4}
\end{figure*}

The proposed SWIN transformer architecture used in the research, exemplified in Figure \ref{fig4}, is a hierarchical vision Transformer designed for various vision tasks. The architecture employs several innovative techniques to process and transform input data efficiently.

\begin{figure}[!htbp]
	\centering
	\includegraphics[width=\columnwidth,height=4cm]{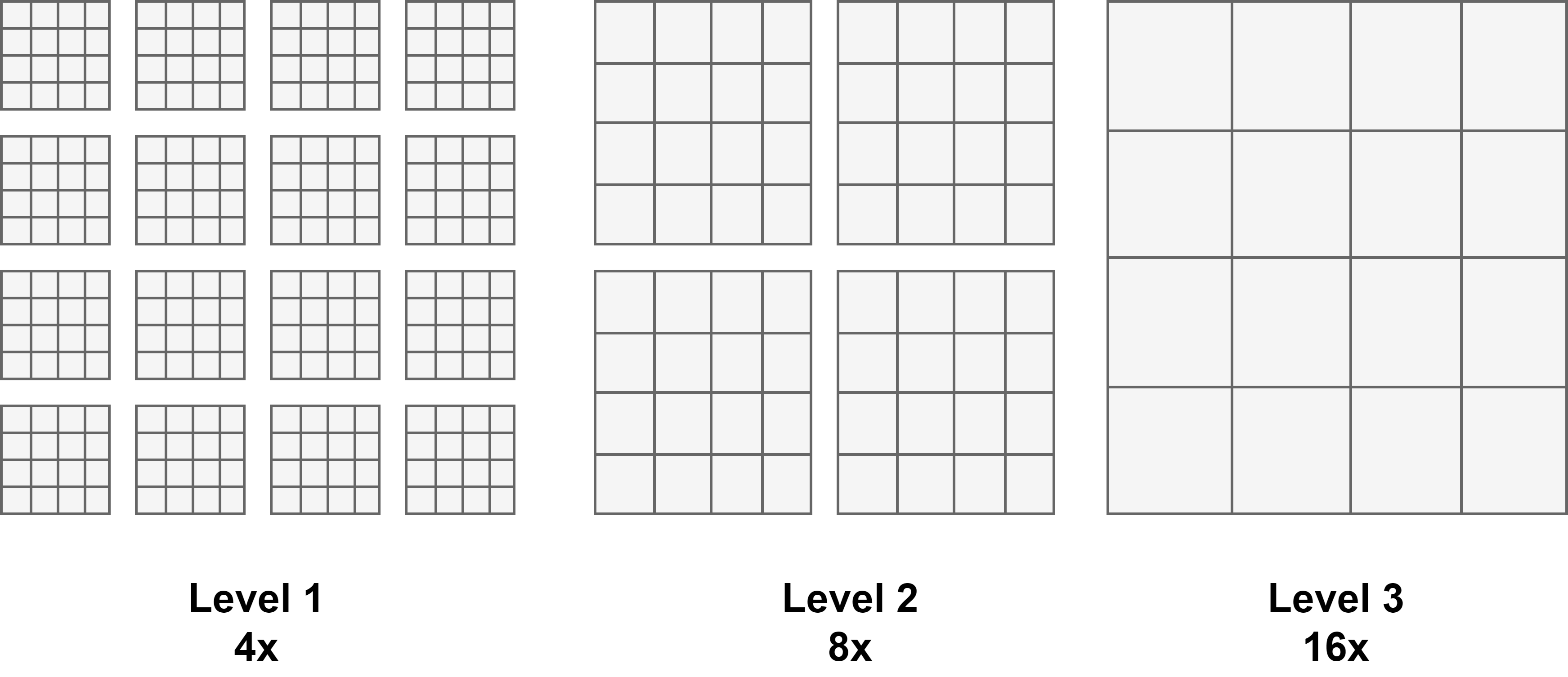}
	\caption{The proposed SWIN Transformer builds a stratified feature map by merging image segmentations in subsequent layers, capturing high and low-resolution details similar to the wavelets concept.
}
	\label{fig5}
\end{figure}

The SWIN Transformer first divides an input image into non-overlapping segments using a patch-splitting module (refer fig. \ref{fig5}), similar to the Vision Transformer (ViT). Each patch is treated as a segmentation, and its features are concatenated through the different layers in the network. For example, given a patch size of $8\times8$, the feature dimension of each segment will be $8\times8\times3 = 192$.
A linear embedding layer then projects these raw features into an arbitrary dimension $C$.

\subsection{Stage 1: Initial Embedding and Transformation}
Several transformer blocks, modified to include SWIN transformer blocks with shifted window-based self-attention module (SW-MSA), applied to these patch feature map segmentations. The initial embedding maintains the number of segmentation is denoted as $\frac{H}{4}\times\frac{W}{4}$. This stage is referred to as Stage 1 as shown in fig. \ref{fig4}.

\subsection{Stage 2: Hierarchical Representation}
The hierarchical representation is achieved by reducing the segmentations by patch-merging layers as the network deepens. The first patch-merging layer concatenates
the features of each group of $2\times2$ neighboring segments and applies a linear layer to the 4C-dimensional concatenated feature maps. This helps reduce the number of segmentations by a factor of a quartet, achieving downsampling the resolution to $2\times 2$. The output dimension is set to $2C$. Following this, the transformer blocks are applied to the feature maps while keeping the resolution at $\frac{H}{8}\times\frac{W}{8}$.

\subsection{Stages 3 and 4: Further Hierarchical Representation}
The procedure is repeated for Stage 3 and Stage 4, with output resolutions of $\frac{H}{16}\times\frac{W}{16}$ and $\frac{H}{32}\times\frac{W}{32}$, respectively. Each stage progressively reduces the number of segmented feature maps and increases the feature dimension, producing a hierarchical representation akin to traditional convolutional networks like VGG and ResNet and vision transformer.

\subsection{Transformer Block}
\begin{figure}[!htbp]
	\centering
	\includegraphics[width=5cm,height=4cm]{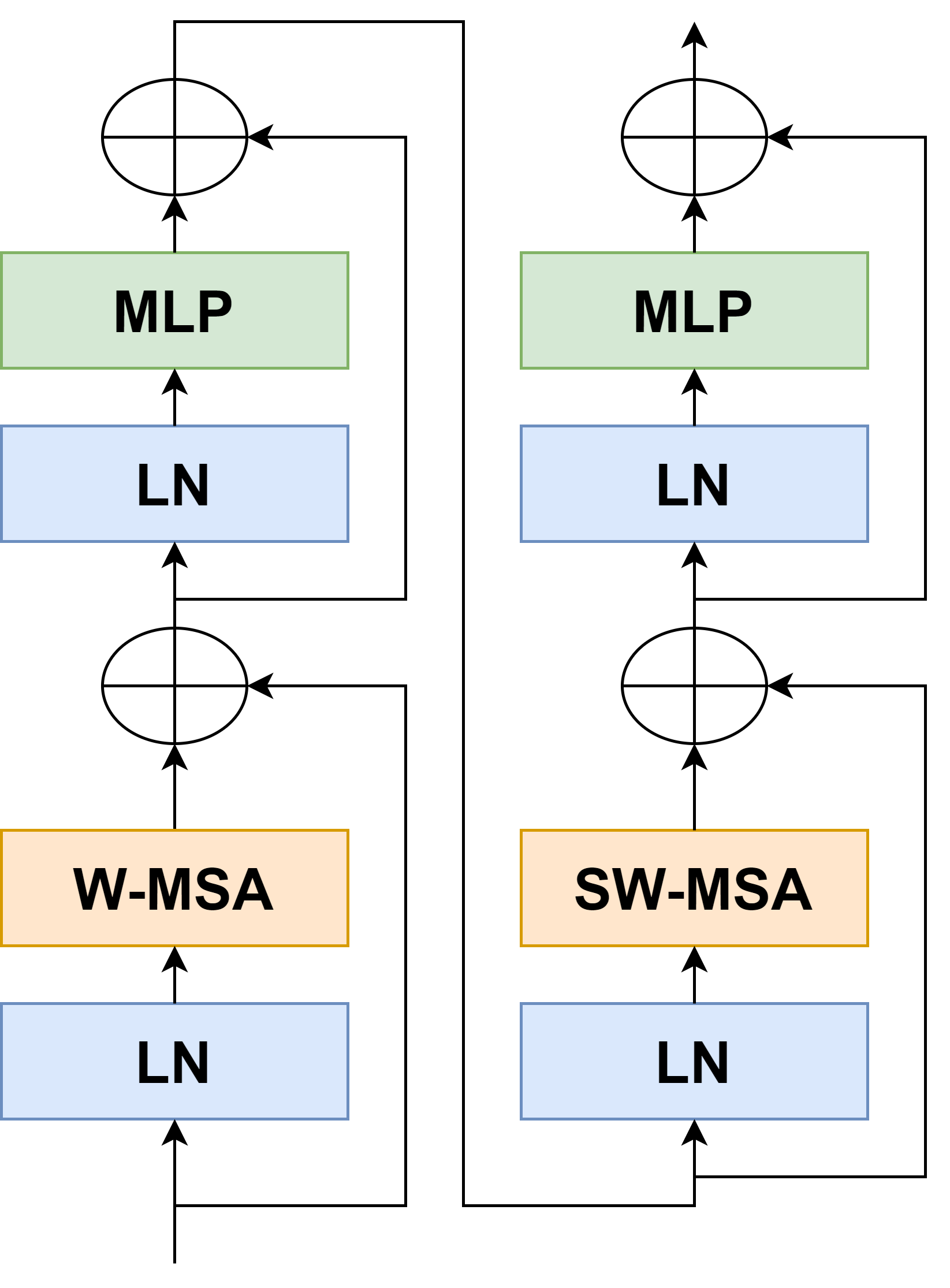}
	\caption{ Block diagram of two successive SWIN transformer Blocks}
	\label{fig 6}
\end{figure}

A SWIN transformer block replaces the standard multi-head self-attention (MSA) module with a shifted window-based MSA module (SW-MSA) blocks in the network stages. For more detail refer to paper by Ze Liu \textit{et al.} \cite{liu2021swin}. Each SWIN transformer block includes the following components (refer fig. \ref{fig 6}:
\begin{itemize}
	\item Shifted window-based MSA module (SW-MSA).
	\item A two-layer multilayer perceptron (MLP) with Gaussian Error Linear Unit (GELU) function.
	\item Normalization layers (LN) applied before each MSA and MLP components.
	\item Residual connections are applied after each module.
\end{itemize}

For the SW-MSA, considering an input feature map dimensions $f \in \mathbb{R}^{H\times W\times C}$. The attention operation within a segmentation can be formulated by Eq \ref{eq4}.

\begin{equation}
	\label{eq4}
	Att(Q,K,V) = SoftMax\left(\frac{QK^T}{\sqrt{d_k}}\right)V
\end{equation}
where, $Q,K$ and $V$ are the query, key and the value matrices, respectively. The dimension of key is $d_k$.
The output of each SWIN transformer block can be expressed by the Eq. \ref{eq5}.
\begin{equation}
	\label{eq5}
	O = MLP(LN(Att(LN(f))))
\end{equation}

\subsection{Classifier and Loss Functions}
Suppose we have $M$ domains $\{\varmathbb{D}^i \mid i = 1:M\}$ and there are two categories in each domain, i.e., output label $y \in \varmathbb{Y} \subset \mathbb{R}$ having value 0/1 representing recaptured/original images. Our objective is to generalize from $\varmathbb{D}$ to unseen target domain $\varmathbb{D}_{M+1}$. The features are generated by the SWIN transformer, for the input images (original and recaptured). To optimize the domain feature generated maps in the backpropagation step, gradient calculation of loss functions (refer sec \ref{preliminary}) are evaluated at the last layer of the network. The classification is done using SoftMax activation function.

\section{Datasets and Experiment Results}
\label{result}

\begin{figure*}
	\centering
	
	\includegraphics[width=15cm]{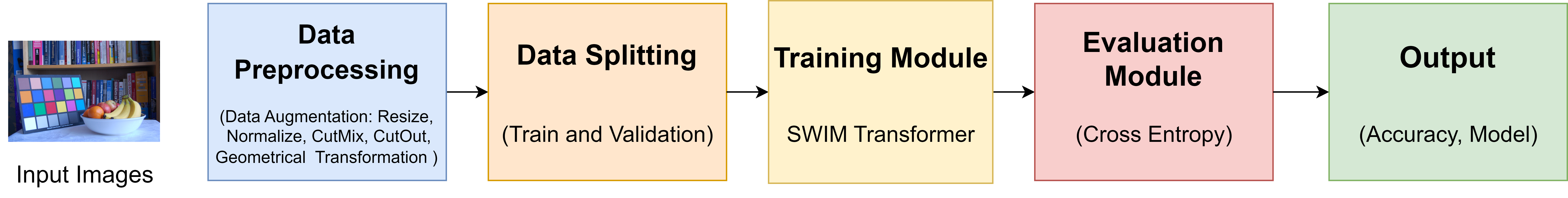}
	\caption{Block diagram of the proposed methodology}
	\label{fig3}
\end{figure*}

The block diagram in Fig. \ref{fig3} illustrates the steps involved in accurately classifying the input image as a recaptured image.  Initially, the input data undergoes preprocessing to eliminate noise caused by varying resolutions and lighting conditions. Subsequently, the dataset is partitioned into training and testing sets. The data is trained using the SWIN Transformer. Finally, the performance of the  model is evaluated considering metrics like accuracy, precision, recall, F1 score, confusion matrix and so on.

\subsection{Datasets}
\label{dataset}
For several reasons, standard datasets are essential in various research fields, particularly in computer vision and machine learning. They provide a common ground for benchmarking and comparing the performance of algorithms, ensuring fair evaluations and reproducibility of research. These datasets contribute to algorithm development and testing, enabling researchers to assess their methods under consistent conditions. Moreover, standard datasets play a vital role in education, providing students with hands-on experience and ethical considerations by ensuring algorithms are tested on diverse and representative datasets. The three publicly available datasets \cite{cao2010statistical,thongkamwitoon2015image,agarwal2018diverse} are present for the LCD recapture image attacks detection. Summarization of the LCD displayed recaptured images from the three datasets (refer to Table \ref{table 3.3}-\ref{table 3.4}). 

\begin{table}[!htbp]
	\centering
	
	\caption{Summary of Publicly Available Datasets for Recaptured Image Detection}
	\label{table 3.3}
	\resizebox{\columnwidth}{!}{
		\begin{tabular}{|l|l|l|l|l|l|}
				\hline
			\multirow{2}{*}{\textbf{Dataset Name}} &
			\multicolumn{2}{c|}{\textbf{Camera 1}} &
			{\textbf{Imitating Medium}} &
			\multicolumn{2}{c|}{\textbf{Camera 2}} \\
			& \centering{\textbf{Model}} & \textbf{Resolution} &  & \textbf{Model} & \textbf{Resolution} \\
				\hline
			
			\multirow{5}{*}{\textbf{NTU-ROSE\footnote[1]{{http://rose1.ntu.edu.sg/datasets/recapturedImages.asp}} }} & - Canon (10D, 400D)  &- 2272$\times$1704 pixels & \textbf{LCD Screens}  &- Canon Powershot & - 2272$\times$1704 pixels\\
			& - Casio & -  4256$\times$2832 pixels&- Philips 19" 190B6CG   & - Olympus Mju&-  4256$\times$2832 pixels\\
			& - Lunix (D1)& & - NEC 17" AccuSync   &- Oylmpus E500 &\\
			&- Nikon (D70, S210)  & &- Acer 17" AL712 &&\\
			&- Sony (Alpha) & & &&\\[1ex]
			\hline

			\multirow{5}{*}{\textbf{ICL\footnote[2]{{http://www.commsp.ee.ic.ac.uk/~pld/research/Rewind/Recapture/}} }} & - Kodak (V550 
			S and B, V610)&- 5MP to 20MP & \textbf{LCD Screens} && - 5MP to 24MP\\
			& - Nikon (D40, D70) & &- NEC MultiSync  IPS  &- Nikon (D3200, D70)&\\
			&- Panasonic (TZ7)&&EA232WMi 23"&- Panasonic (TZ7, TZ10)&\\
			&- Canon (600D)&&&- Canon (60D, 600D)&\\
			&- Olympus (E-PM2)&&&- Olympus (E-PM2)&\\
			&- Sony (RX100)&&&- Sony (RX100)&\\
			\hline
			
			\multirow{9}{*}{\textbf{Mturk\footnote[3]{{https://agarwalshruti15.github.io/}}}} & - Apple Iphone  & 5MP to 20 MP& \textbf{LCD Screens}& - Nikon& - 5MP to 20 MP\\
			& - Canon &&&- Samsung&\\
			& - Casio & &\textbf{Scanners}&- Fujifilm&\\
			&- Fujifilm &&&- Apple Iphone&\\
			&- Kodak &&\textbf{Printers}&- Canon&\\
			&- Leica &&&- Sony&\\
			&- Parasonic &&\textbf{Screengrab}&- Panasonic&\\
			&- Nikon &&&- HTC&\\
			&- Samsung &&&- Kotak&\\
			
			\hline
	\end{tabular}}
\end{table}
\begin{table}[!htbp]
	\centering
	\caption{Attributes summarization for the three datasets}
	\label{table 3.4}
	\resizebox{\columnwidth}{!}{%
		%{\begin{tabular*}{}{@{\extracolsep{\fill}}lccc@{}}
		%	\label{table 2}
		\begin{tabular}{|l|c|c|c|}
			\hline
			
			\textbf{Attributes} & \textbf{NTU-ROSE Dataset}\cite{cao2010statistical} & \textbf{ICL Dataset}\cite{thongkamwitoon2015image} & \textbf{Mturk Dataset}\cite{agarwal2018diverse} \\ \hline
			\textbf{Year} & 2010 & 2015 & 2018 \\ \hline
			\textbf{Number of Original Images} & 2710 & 900 & 3956 \\ \hline
			\textbf{Number of Recaptured Images} & 2776 & 1400 & 3873 \\ \hline
			\textbf{LCD Screen Count} & 3 & 1 & 129 \\ \hline
			\textbf{Original Camera Quantity} & 5 & 8 & 1036 \\ \hline
			\textbf{Reacquisition Camera Quantity} & 3 & 9 & 119 \\ \hline
			\textbf{Alias-Free Status} & No & Yes & No \\ \hline
			\textbf{Presence of Tampered Content} & Yes & No & No \\ \hline
			\textbf{Original Image Format} & JPEG & JPEG & JPEG \\ \hline
			\textbf{Recaptured Image Format} & JPEG & PNG & PNG \\ \hline
	\end{tabular}}
\end{table}

\begin{figure}[!htbp]
	\centering
	\subfloat[\label{fig 3.7a}]{\includegraphics[height=2cm,width=2cm]{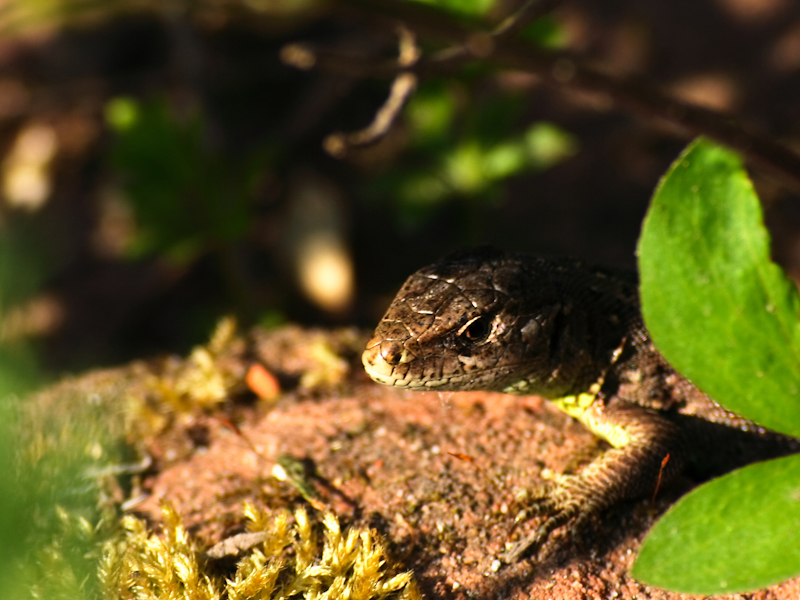}}\quad
	\subfloat[\label{fig 3.7b}]{\includegraphics[height=2cm,width=2cm]{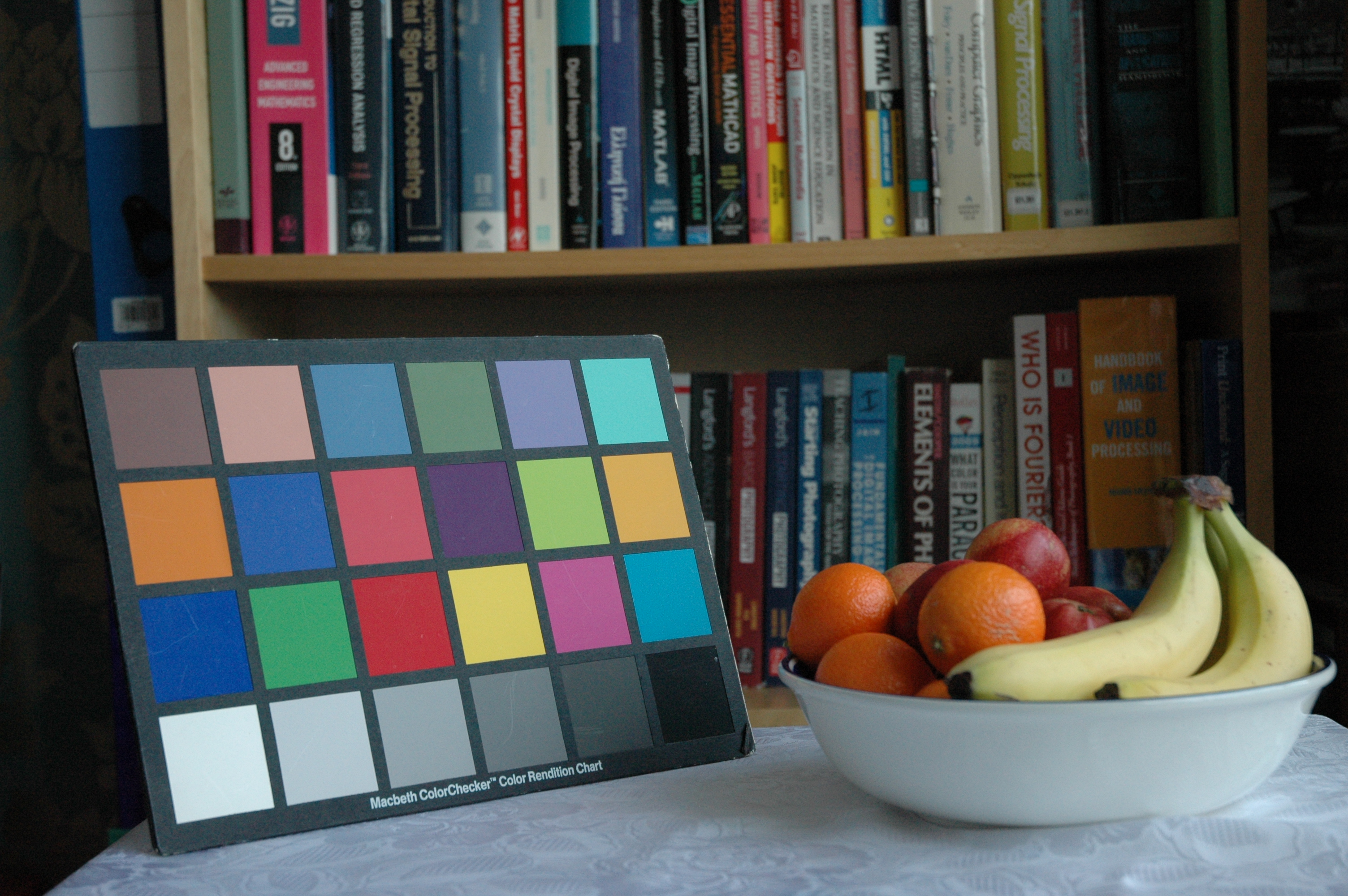}}\quad
	\subfloat[\label{fig 3.7c}]{\includegraphics[height=2cm,width=2cm]{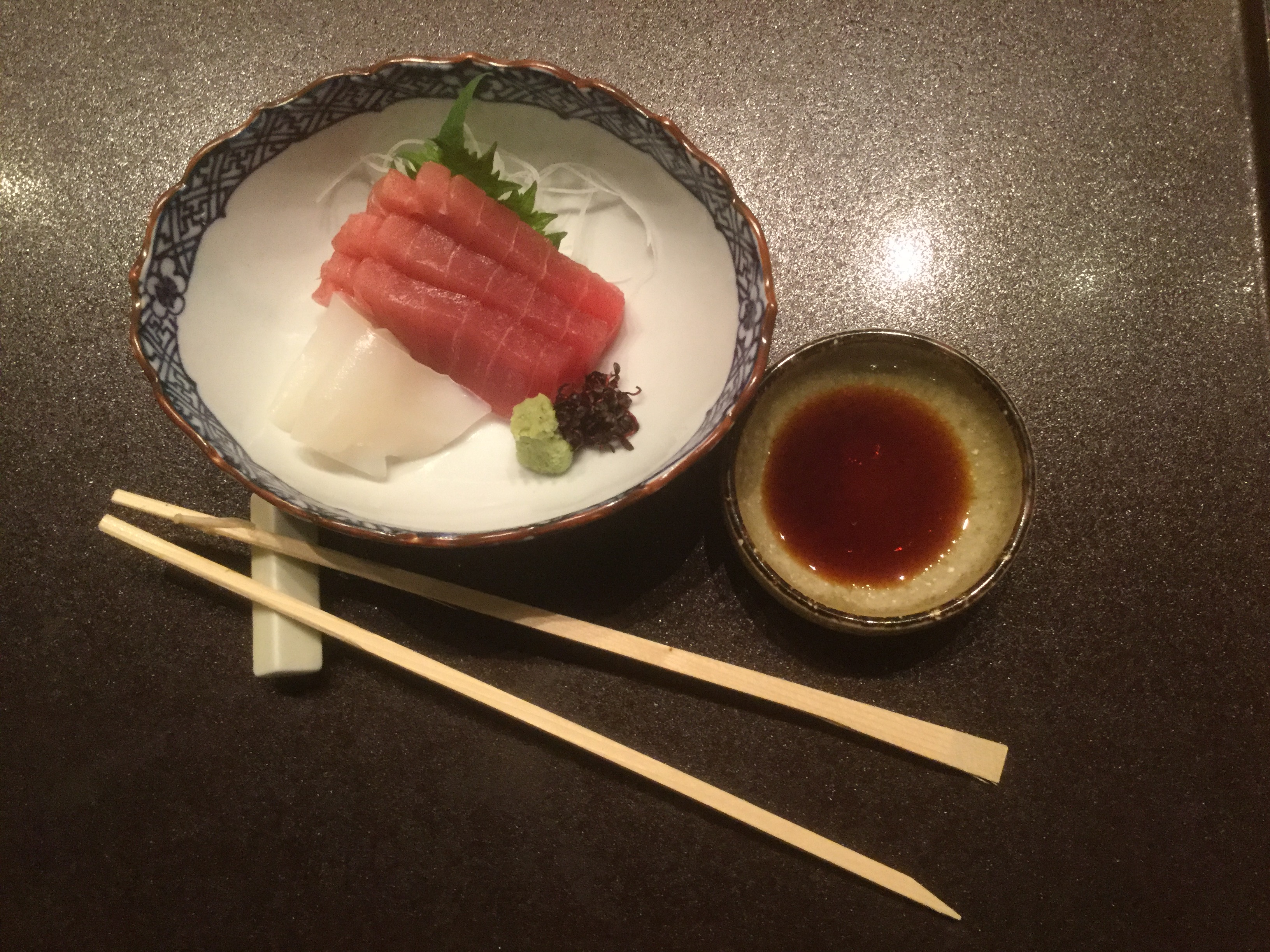}}\quad
	%\subfloat[\label{fig 3.7d}]{\includegraphics[height=2cm,width=2cm]{ck_orig4.png}}\\
	\subfloat[\label{fig 3.7e}]{\includegraphics[height=2cm,width=2cm]{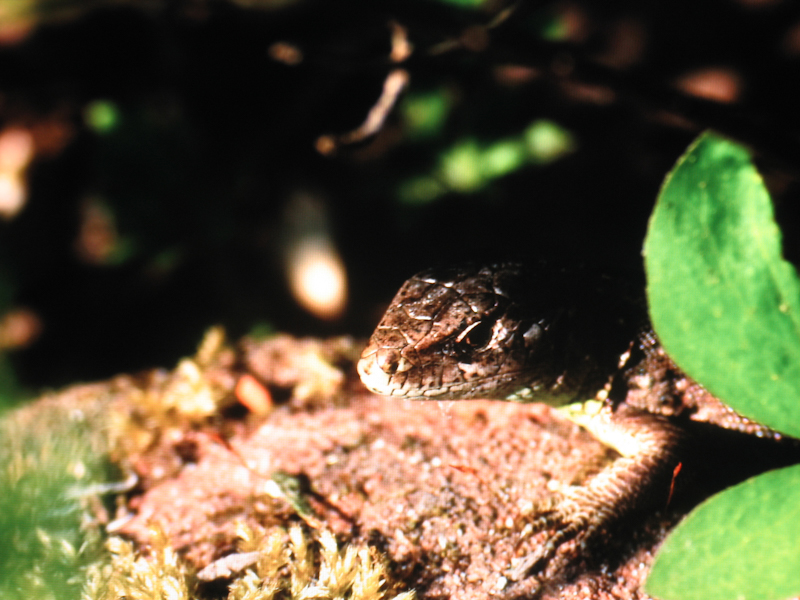}}\quad
	\subfloat[\label{fig 3.7f}]{\includegraphics[height=2cm,width=2cm]{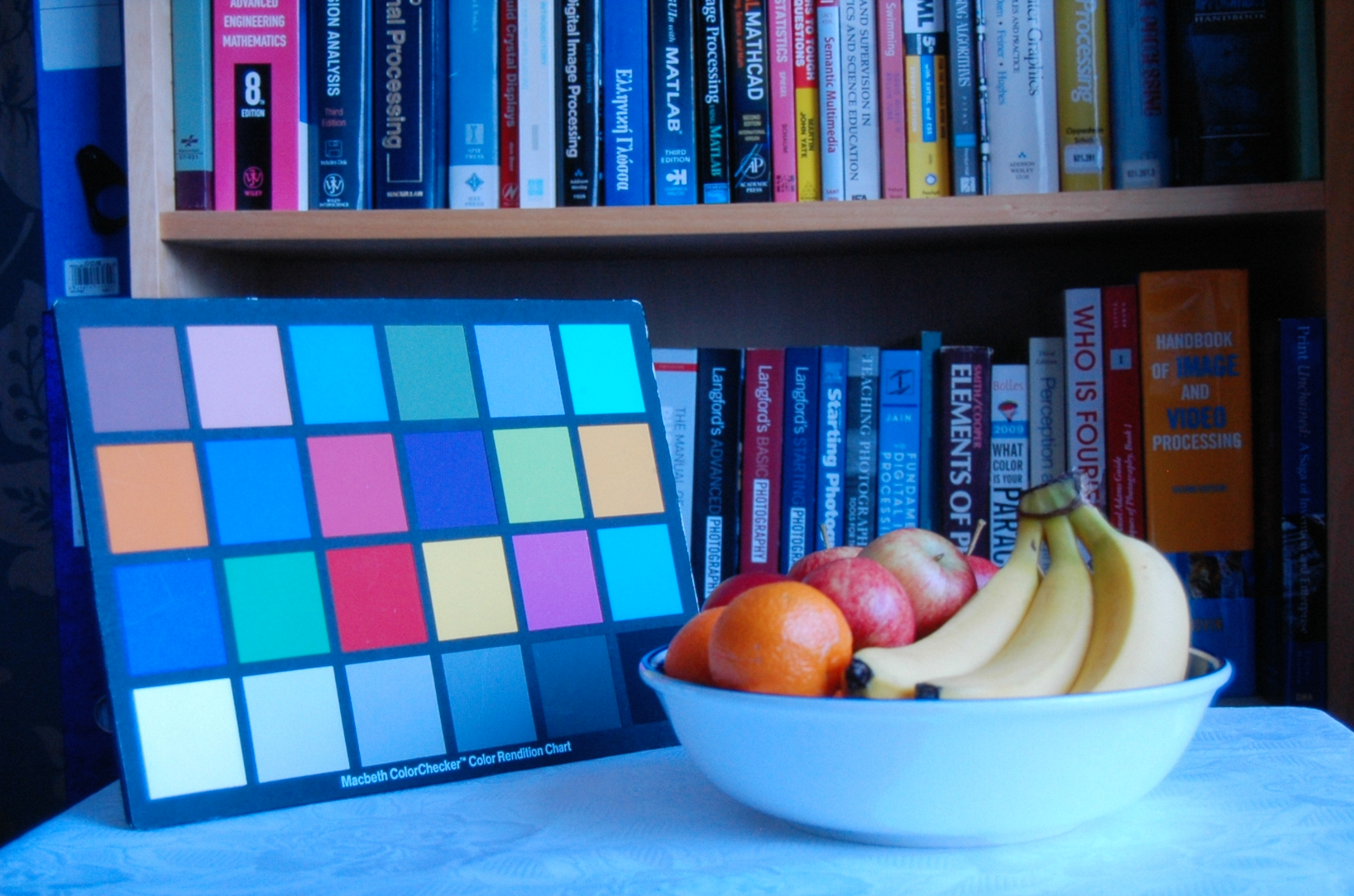}}\quad
	\subfloat[\label{fig 3.7g}]{\includegraphics[height=2cm,width=2cm]{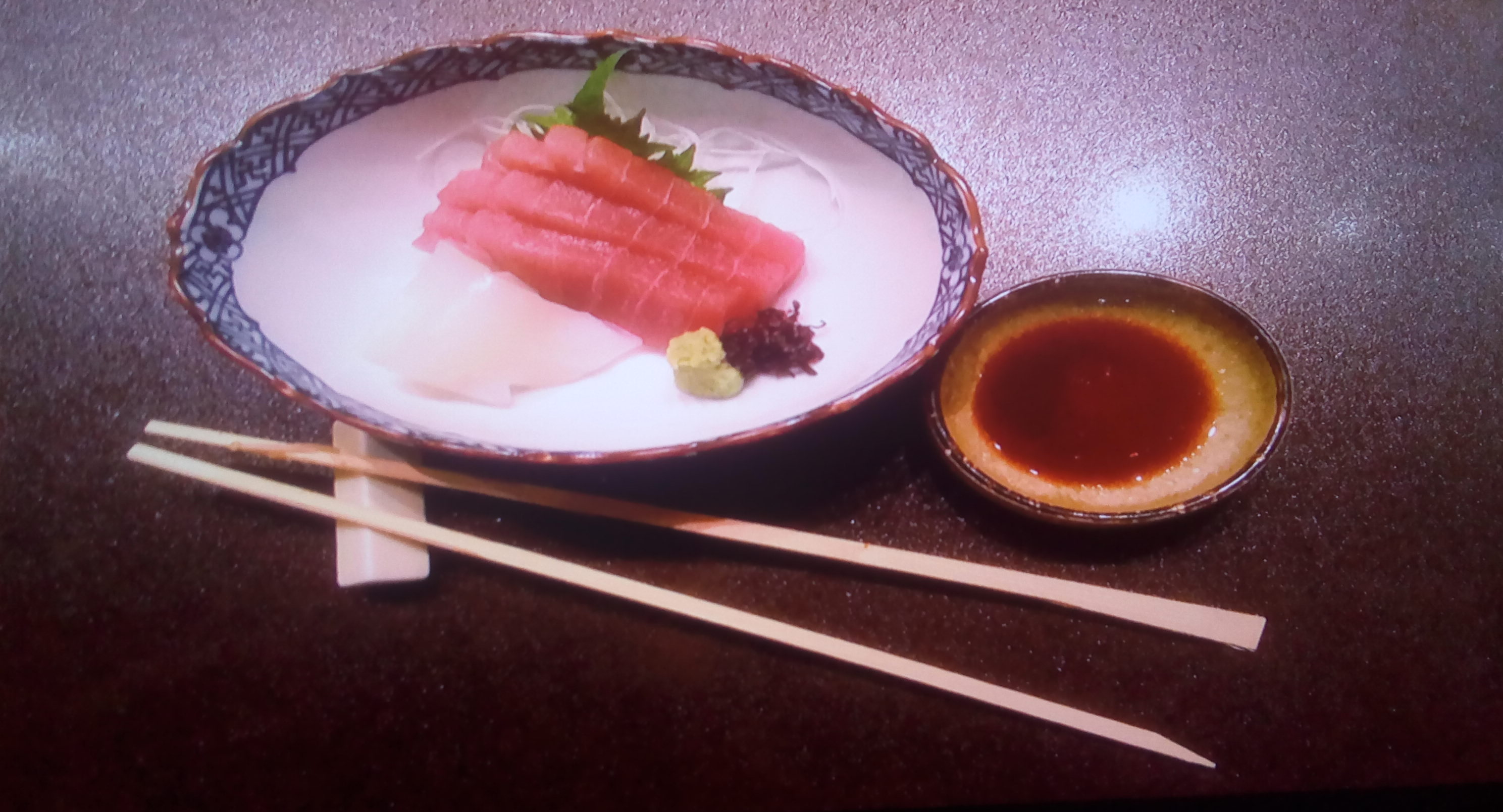}}\quad
	%\subfloat[\label{fig 3.7h}]{\includegraphics[height=2cm,width=2cm]{ck_recap4.png}}
	\caption{The provided imagery showcases three singly captured images sourced from the NTU-ROSE, ICL and Mturk databases (left to right) in the top row, alongside their corresponding recaptured counterparts in the bottom row, revealing the evident visual correspondence between all three domains}
	
	\label{Fig 3.7}
\end{figure}

\begin{figure}[!htbp]
	\centering
	\includegraphics[width=5cm,height=4cm]{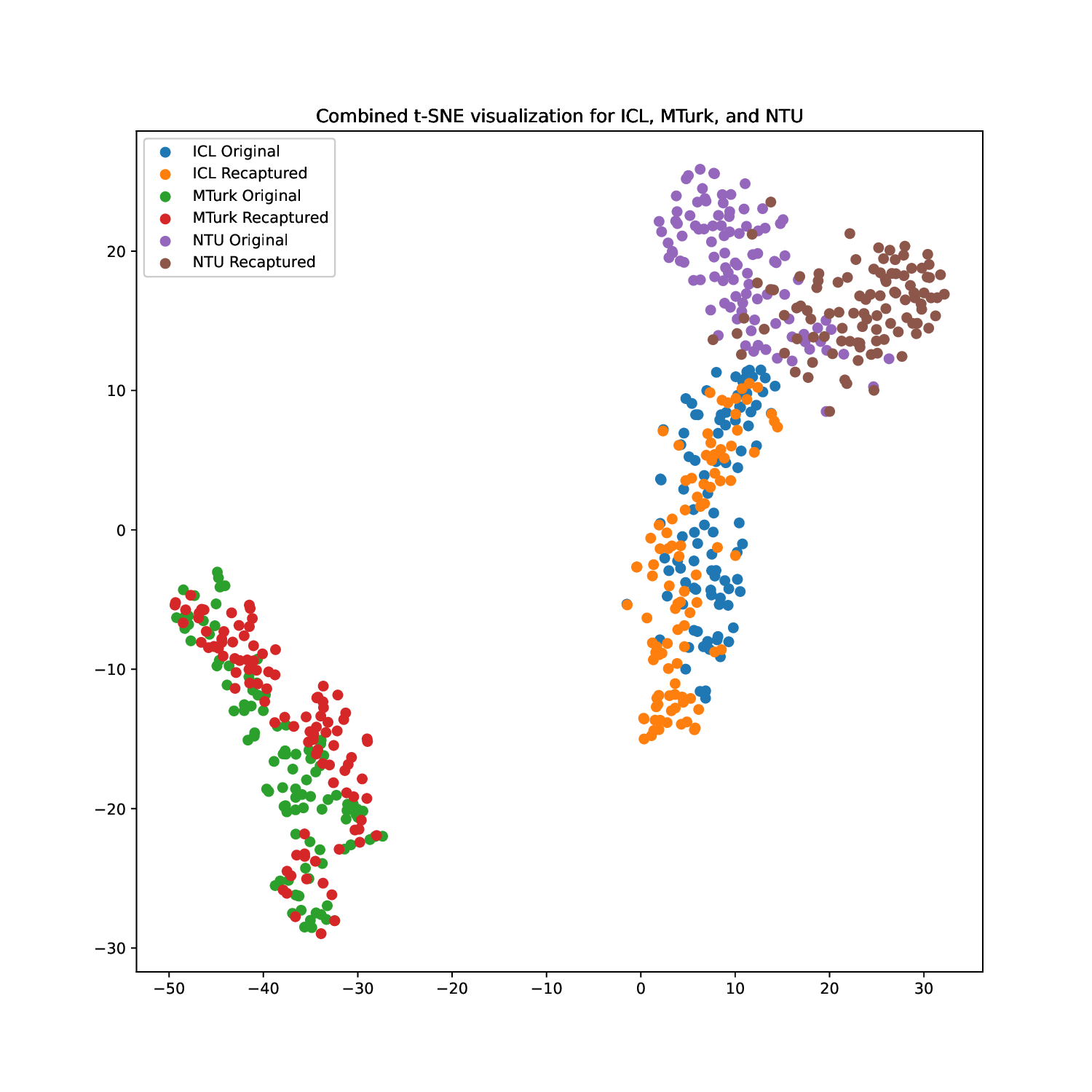}
	\caption{The t-SNE visualization plot of the extracted features from 200 sample images from each datasets: D1, D2 and D3 using the proposed SWIN transformer model.}
	\label{fig7}
\end{figure}

Examples from the NTU-ROSE, ICL and Mturk datasets are illustrated in Fig. \ref{Fig 3.7}. It can
be observed by comparing both classes of images from different datasets that the recapturing process introduces artefacts such as noise, blurring and color distortion. To show the variations in the resultant features embedding among the three datasets sampled images for the two classes, we have ploted the two dimension t-SNE visualization plot (refer fig. \ref{fig7}).

There are a few critical points to be noted. The red and green dots represent the features extracted from the recaptured and original images from the D3 dataset. The brown and purple dots represent the D1 dataset features, and the orange and blue dots represent the D2 dataset images. First observation, the captured and recaptured samples from D1 and D2 are clustered together. However, they are collected by two sets of different devices and different capturing environments but have high-resolution images. It demonstrates the difficulty in the generalization performance of the proposed deep model using these images for training and testing it for the D3 dataset because of the difference between the training and testing features. Second,  most of the samples from D3 are found to be in a single cluster. This indicates that the proposed scheme acceptably distinguishes between D3 and D1/D2 images. Third, this follows for both classes of images from all three datasets.

\subsection{Experimention Settings and Hyperparameters}
In experimental setup, we employ our DAST network to extract features. The parameters of each module in a network are explained in the subsection \ref{methodology}. We have made our code publicly available on \textbf{\textup{Github URL}}. Adam optimizer is used to automatically adjust the learning rate and fine-tune our model for 10 epochs. In addition, our model is implemented in PyTorch, running on a Dell inspiron 5502 with CPU processor i5 and RAM of 16GB. The details of hyperparameters is mentioned in table \ref{table5}. The evaluation parameters used are accuracy, precision, recall values and AUC (Area Under the receiver operating characteristic Curve) plots for the testing datasets. The value of AUC provides the area under the ROC curve. The higher value indicates a better authentication performance for the recaptured image classification. Also, we have the accuracy and loss plots generated for 10 epochs showing the training and validation results.

\begin{table}[!htbp]
	\centering
	\caption{Hperparameters}
	\label{table5}
	\begin{tabular}{|l|c|}
		\hline
		\textbf{Hyperparameters} & \textbf{Values}\\
		\hline
		Learning Rate & 0.0001\\ \hline
		Optimizer & Adam\\ \hline
		Loss functions &Cross Entropy\\ \hline
		Batch Size& 32\\ \hline
		Input Image Size &224$\times$224$\times$3\\ \hline
		Normalized Mean&[0.485, 0.456, 0.406]\\ \hline
		Normalized Std &[0.229, 0.224, 0.225]\\ 
		\hline
	\end{tabular}
\end{table}

\subsection{Experimental Results and Discussion}
To investigate the challenges posed by LCD recaptured images, we compare the performances of the proposed technique with some state-of-the-art methods on the three databases under different experimental settings. The table \ref{table6} shows the testing and training datasets combinations.

\begin{table}[!htbp]
	\centering
	\caption{Various Experiment settings. D1: NTU-ROSE dataset, D2: ICL dataset, and D3: Mturk dataset}
	\label{table6}
	\begin{tabular}{|l|l|l|}
		\hline
		\textbf{Experiment Type} & \textbf{Training Datasets} & \textbf{Testing Datasets}\\
		\hline
		\multirow{3}{*}{\textbf{Intra-Domain}} & D1 & D1\\ \cline{2-3}
		&D2&D2\\ \cline{2-3}
		&D3&D3\\ \hline
		\textbf{Inter-Domain}&D1+D2+D3& D1+D2+D3\\
		\hline
		\multirow{3}{*}{\textbf{Cross-Domain}} & D1+D2 & D3\\ \cline{2-3}
		&D2+D3&D1\\ \cline{2-3}
		&D3+D1&D2\\
		\hline
	\end{tabular}
\end{table}

\subsubsection{Intra and Inter-Domain Results} 
\begin{figure}[!htbp]
\centering
\subfloat[\label{fig9a}]{\includegraphics[trim={13cm 0 0 0},clip,width=3cm,height=3cm]{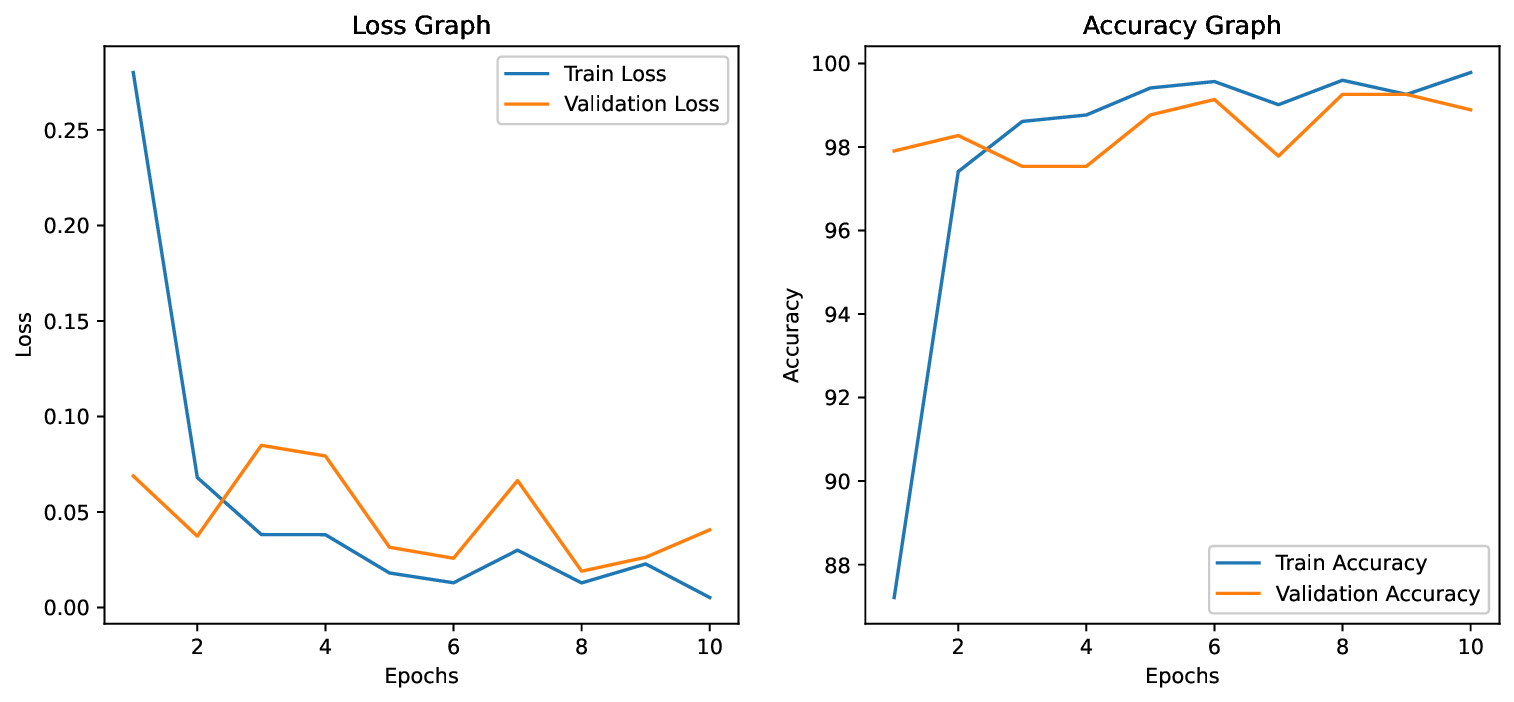}}
\subfloat[\label{fig9b}]{\includegraphics[width=3cm,height=3cm]{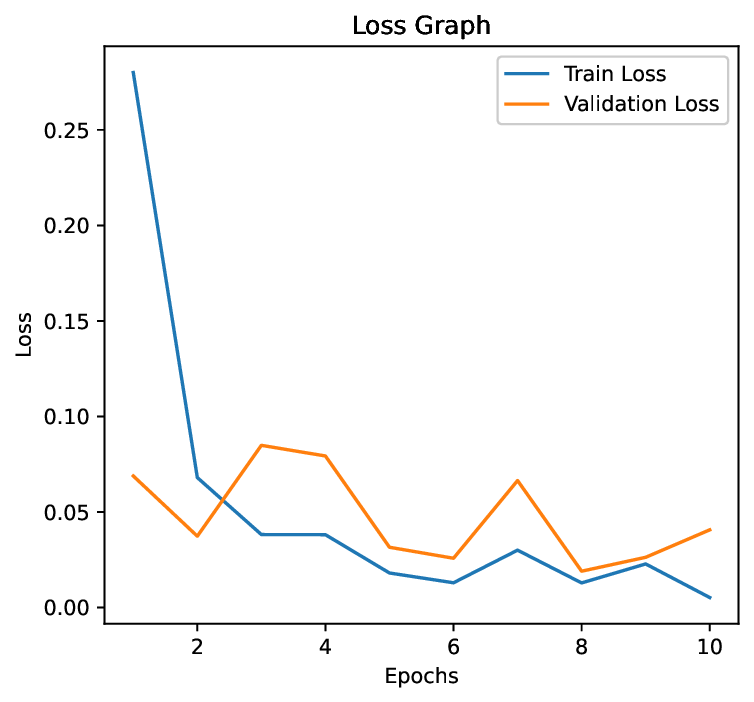}}
\subfloat[\label{fig9c}]{\includegraphics[width=3cm,height=3cm]{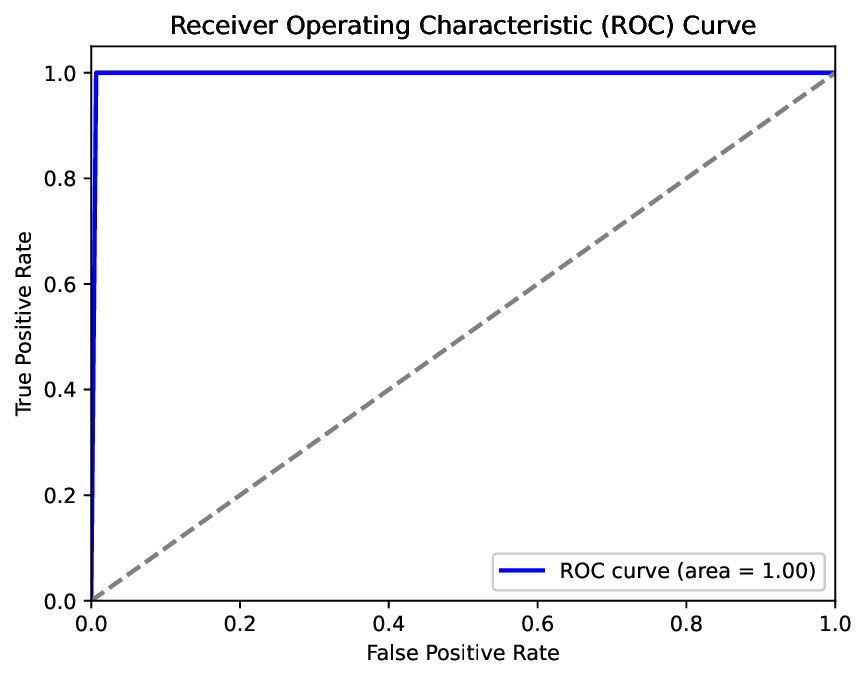}}\\

\subfloat[\label{fig9d}]{\includegraphics[width=3cm,height=3cm]{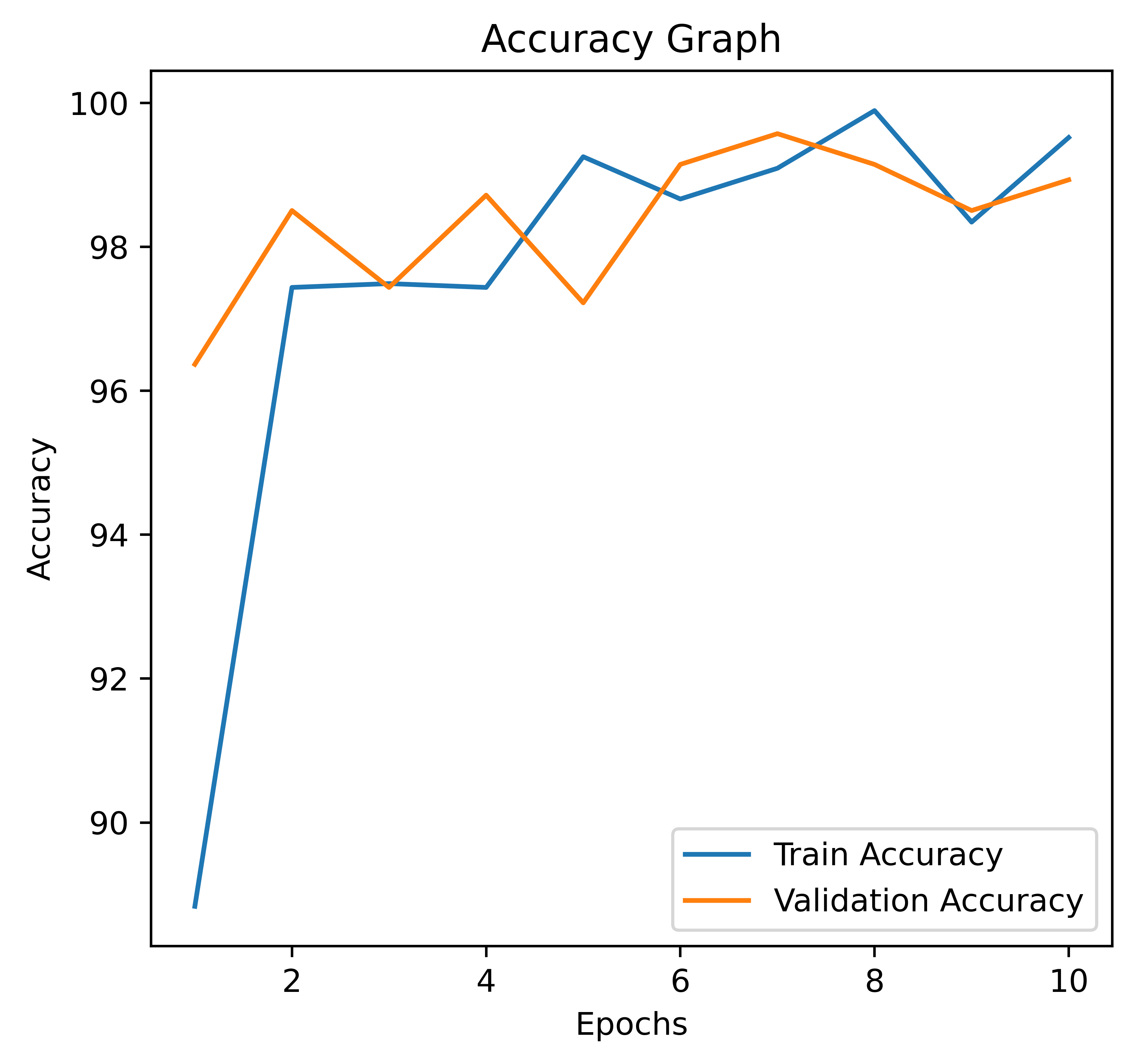}}
\subfloat[\label{fig9e}]{\includegraphics[width=3cm,height=3cm]{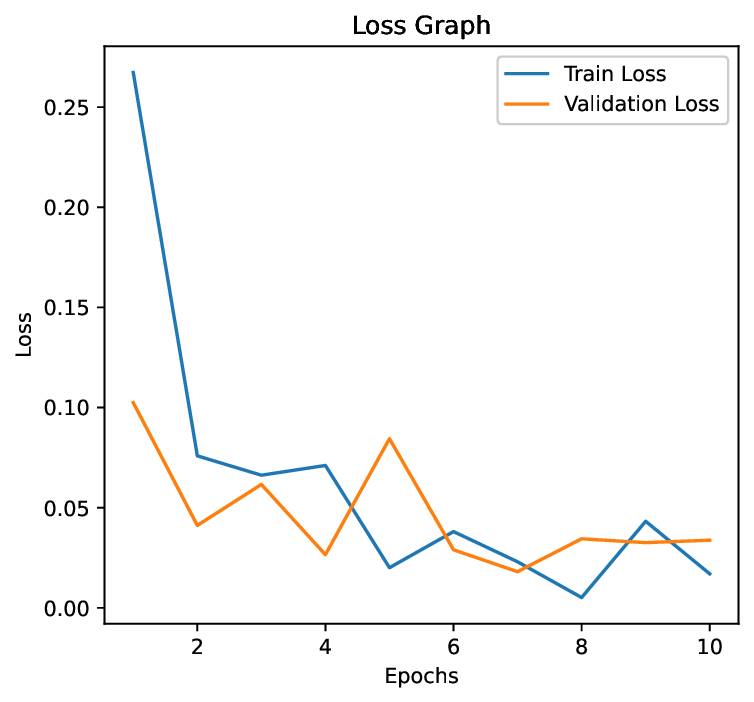}}
\subfloat[\label{fig9f}]{\includegraphics[width=3cm,height=3cm]{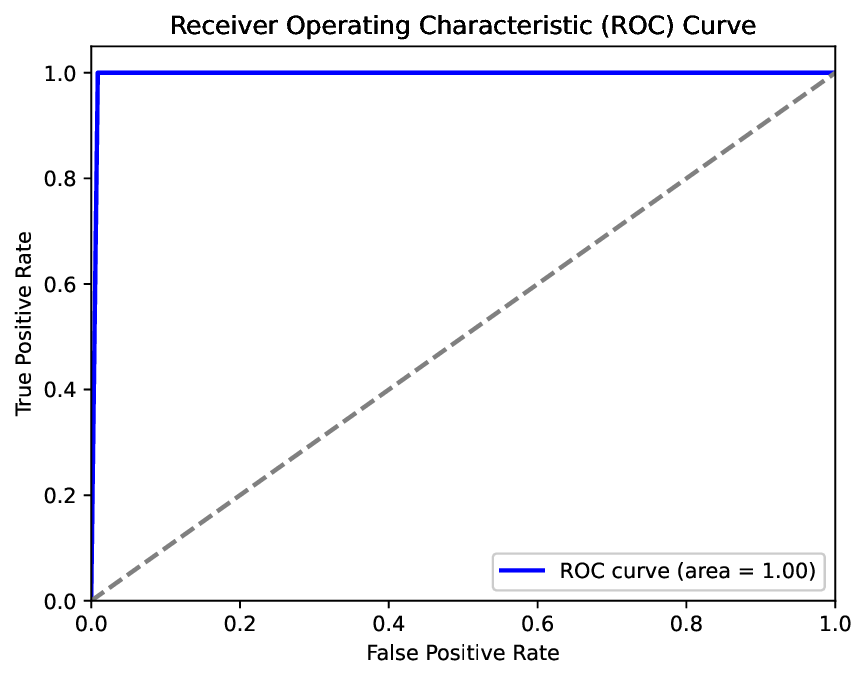}}\\

\subfloat[\label{fig9g}]{\includegraphics[width=3cm,height=3cm]{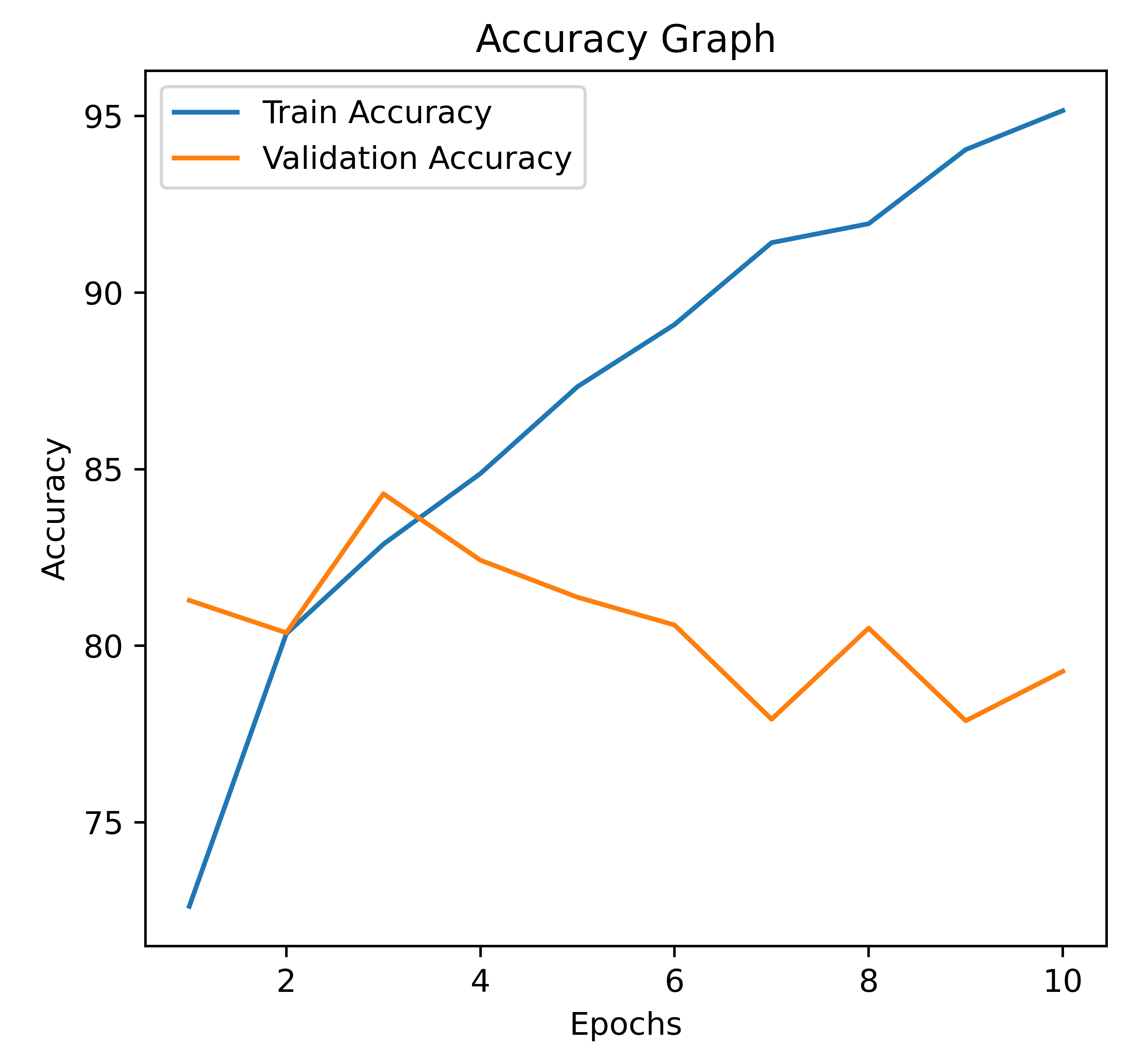}}
\subfloat[\label{fig9h}]{\includegraphics[width=3cm,height=3cm]{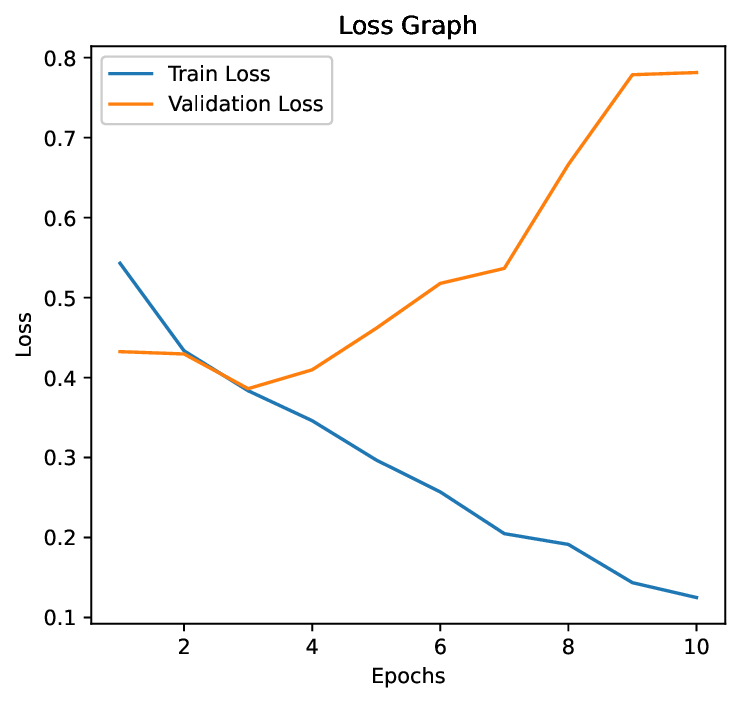}}
\subfloat[\label{fig9i}]{\includegraphics[width=3cm,height=3cm]{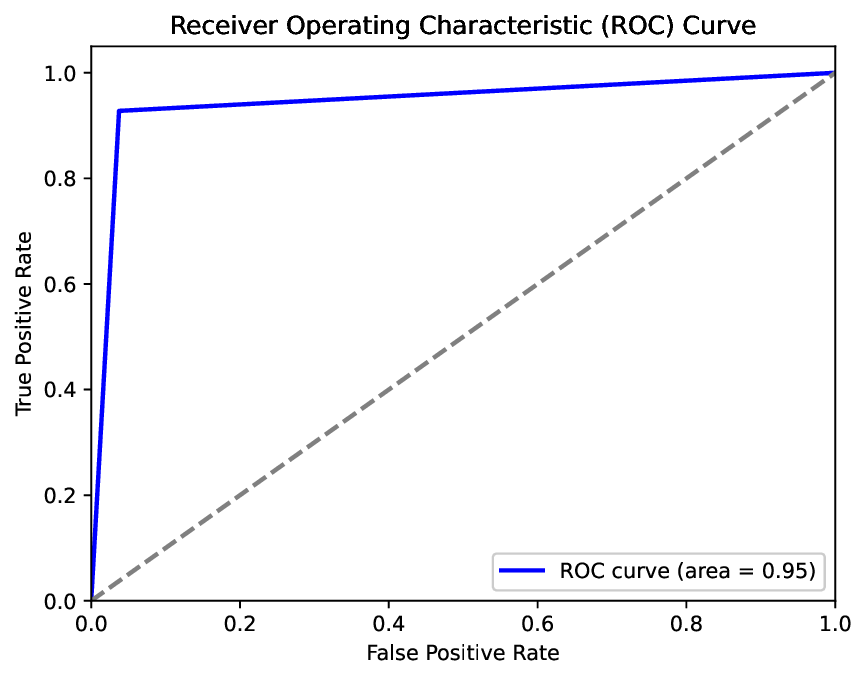}}\\

\subfloat[\label{fig9j}]{\includegraphics[trim={13cm 0 0 0},clip,width=3cm,height=3cm]{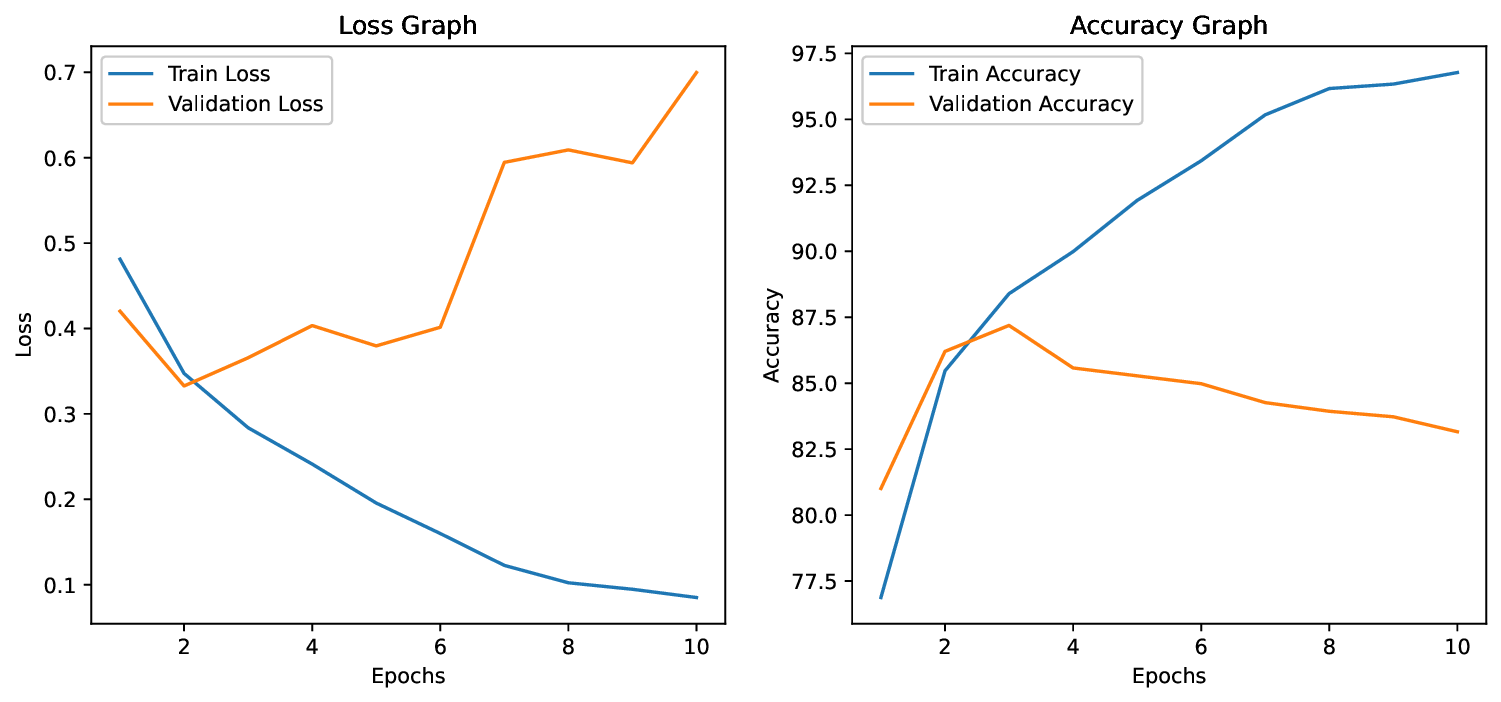}}
\subfloat[\label{fig9k}]{\includegraphics[width=3cm,height=3cm]{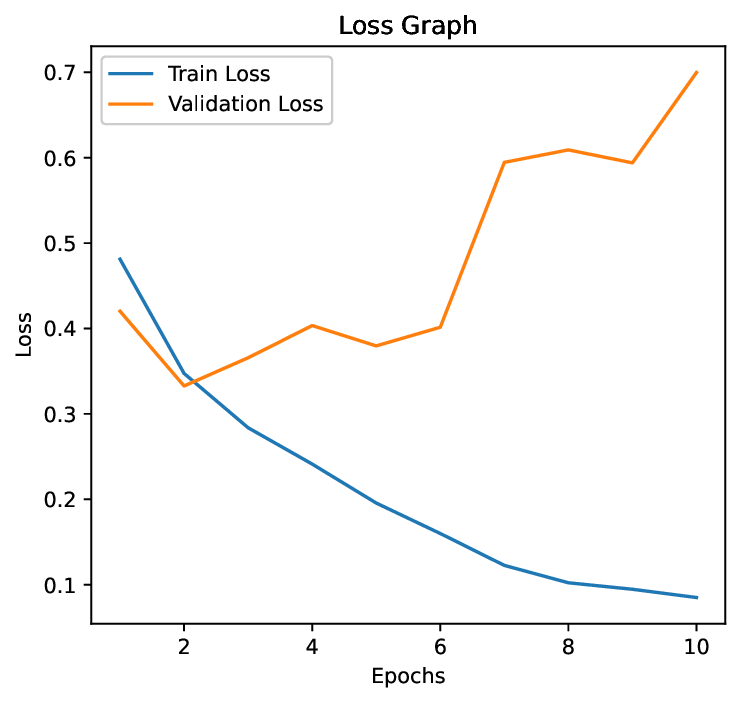}}
\subfloat[\label{fig9l}]{\includegraphics[width=3cm,height=3cm]{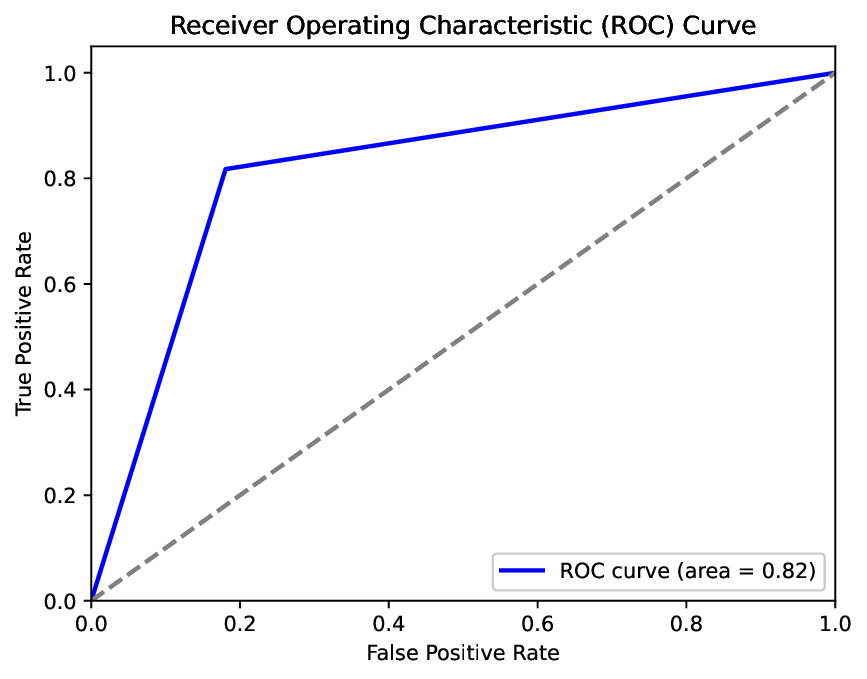}}

\caption{From top to bottom row, representing the results of NTU-ROSE (D1), ICL (D2), Mturk (D3) and combined (D1+D2+D3) datasets, respectively. From left to right, training and validation accuracy plot, loss plot and testing ROC plot, respectively generated for ten epochs.}
\label{fig9}

\end{figure}

The training and testing images are from the same dataset acquired by the different devices but under the same environmental conditions. Also, the training and testing datasets have additional sample images from data augmentation. The data augmentation helps generalize the model and avoids the over-fitting issue. The loss function used is cross-entropy with a softmax classifier. For the intra-database experiment, the results show that the AUCs of our proposed model are 99.77\%, 99.65\%, and 94.53\% for datasets I, II, and III, respectively. 
For the inter-domain experiment, the AUCs is 81.84\%. The AUC value is dropped for the inter-domain compared to the intra-domain testing (refer table \ref{table7}). The training and validation accuracy, loss, and ROC curves for ten epochs are shown in Fig. \ref{fig9}. Also, from the t-SNE, we can observe that the D1 dataset has distinguished features for the original and recaptured classes, but the remaining two datasets have no such distinguishment for classification. This is reflected in the precision and recall values for all three datasets.

\begin{table}[!htbp]
	\centering
	\caption{Experimental results for the datasets D1, D2 and D3. The datasets were divided in an 8:1:1 ratio for the training, validation and testing. All the evaluation parameters are in \% except the samples which are the count value of images in the training and testing phase.}
	\label{table7}
	\resizebox{\columnwidth}{!}{
	\begin{tabular}{|l|l|l|l|l|l|l|}
		\hline
		\textbf{Datasets} & \textbf{Training Samples} & \textbf{Testing Samples} & \textbf{Accuracy} & \textbf{Precision} & \textbf{Recall} & \textbf{F1-score} \\ \hline
		\textbf{D1} & 3245 & 4057 & 99.77 & 99.66 & 100 & 99.83 \\ \hline
		\textbf{D2} & 1872     & 2340     &  99.65     &    99.44   & 100    & 99.72      \\ \hline
		\textbf{D3} &   9145   &  11432    &   94.53    &  96.05     &  92.78   &    94.38   \\ \hline
		\textbf{D1+D2+D3}&   10418   &   3473   &    81.84   &    82.33   &   81.75  &   82.04    \\ \hline
	\end{tabular}}
\end{table}

\subsubsection{Cross-dataset Experiment} 
\begin{figure}[!htbp]
	\centering
	\subfloat[\label{fig10a}]{\includegraphics[trim={13cm 0 0 0},clip,width=3cm,height=3cm]{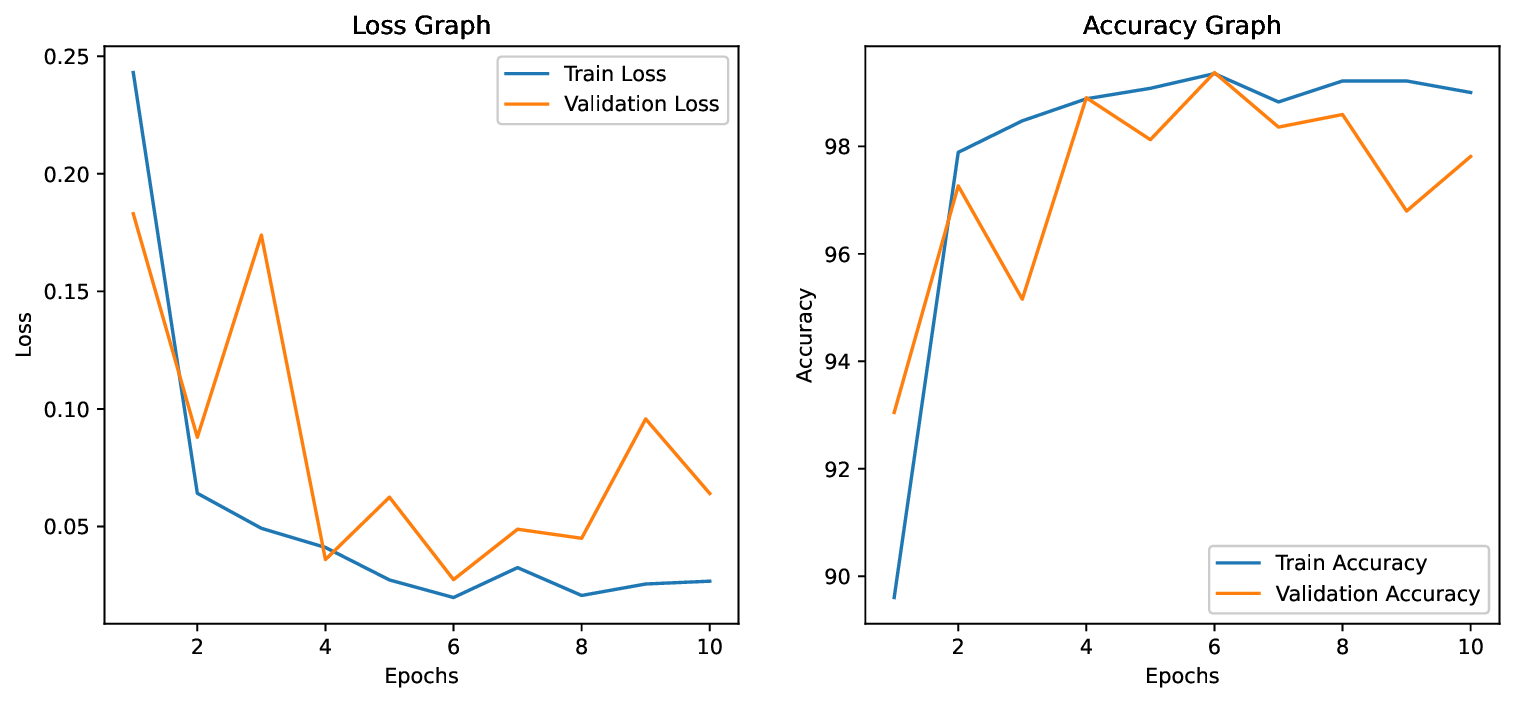}}
	\subfloat[\label{fig10b}]{\includegraphics[width=3cm,height=3cm]{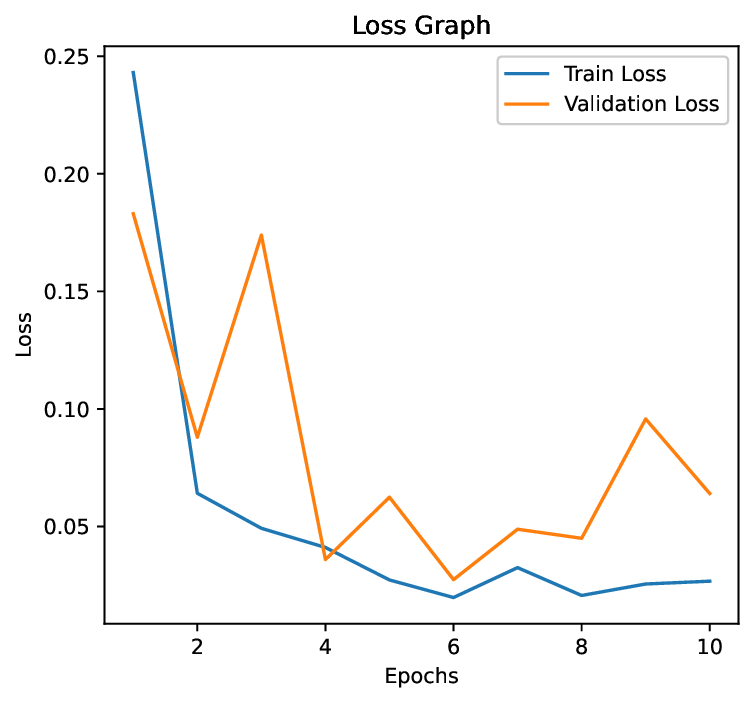}}
	\subfloat[\label{fig10c}]{\includegraphics[width=3cm,height=3cm]{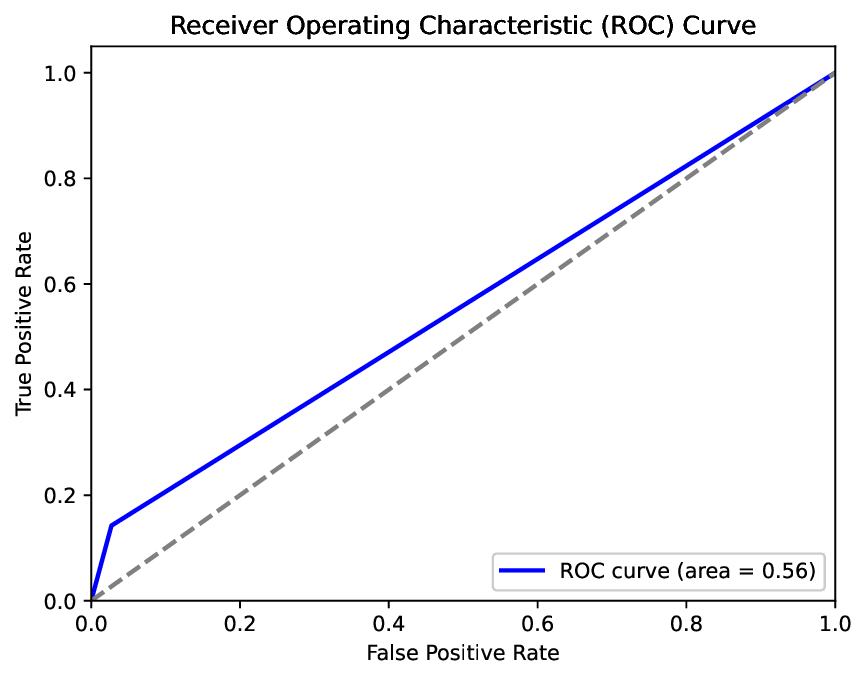}}\\
	
	\subfloat[\label{fig10d}]{\includegraphics[trim={13cm 0 0 0},clip,width=3cm,height=3cm]{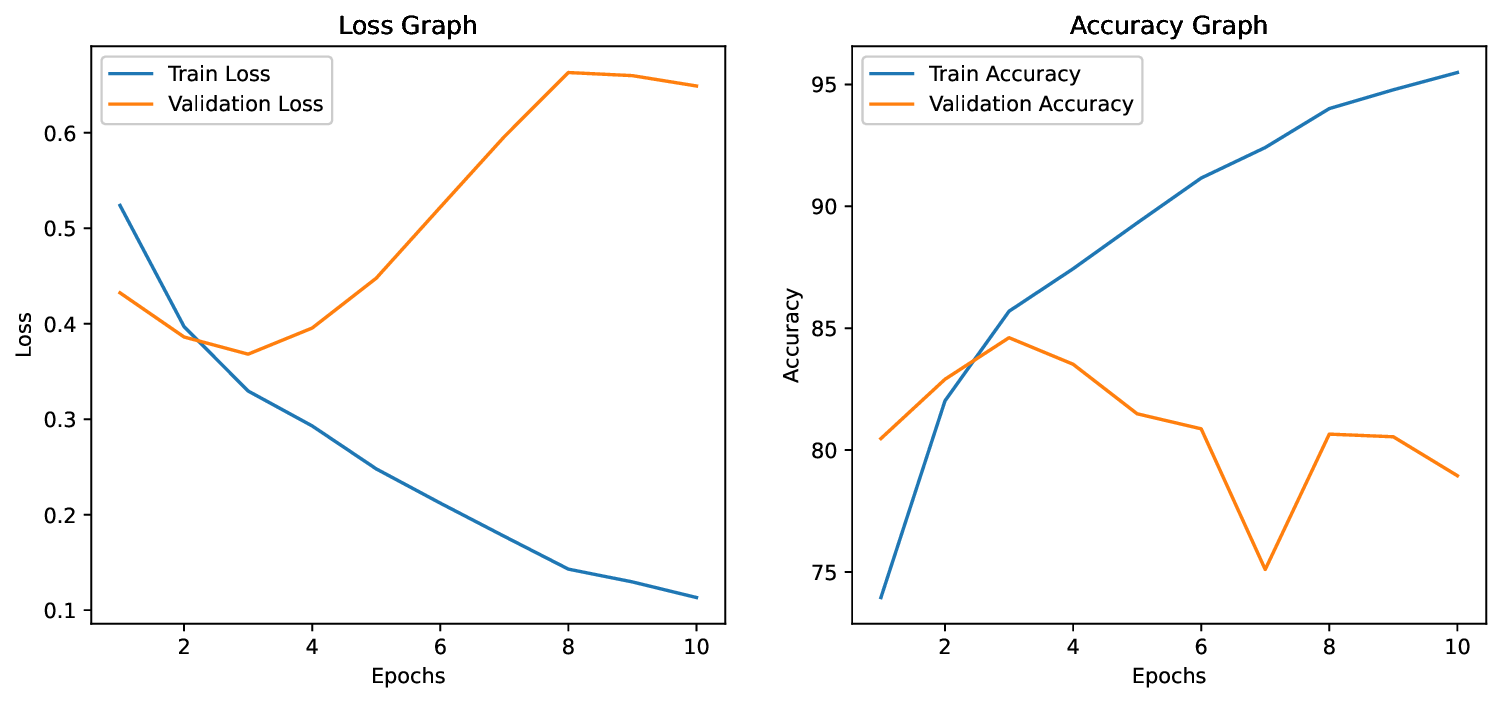}}
	\subfloat[\label{fig10e}]{\includegraphics[width=3cm,height=3cm]{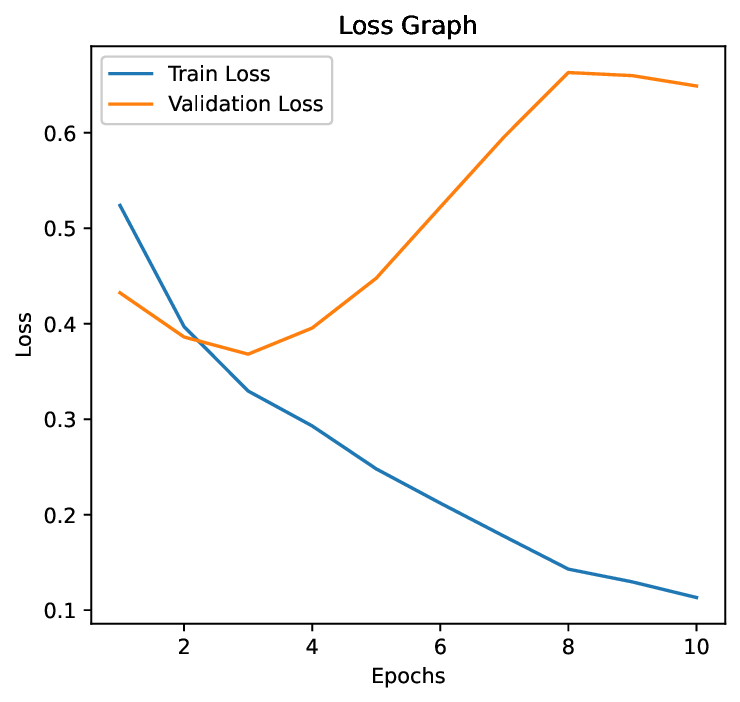}}
	\subfloat[\label{fig10f}]{\includegraphics[width=3cm,height=3cm]{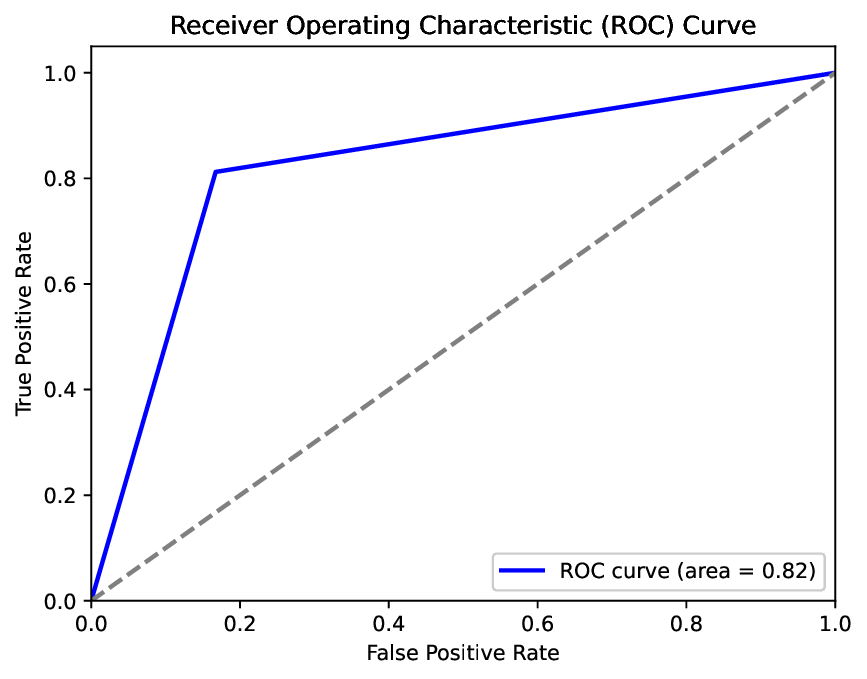}}\\
	
		\subfloat[\label{fig10g}]{\includegraphics[trim={13cm 0 0 0},clip,width=3cm,height=3cm]{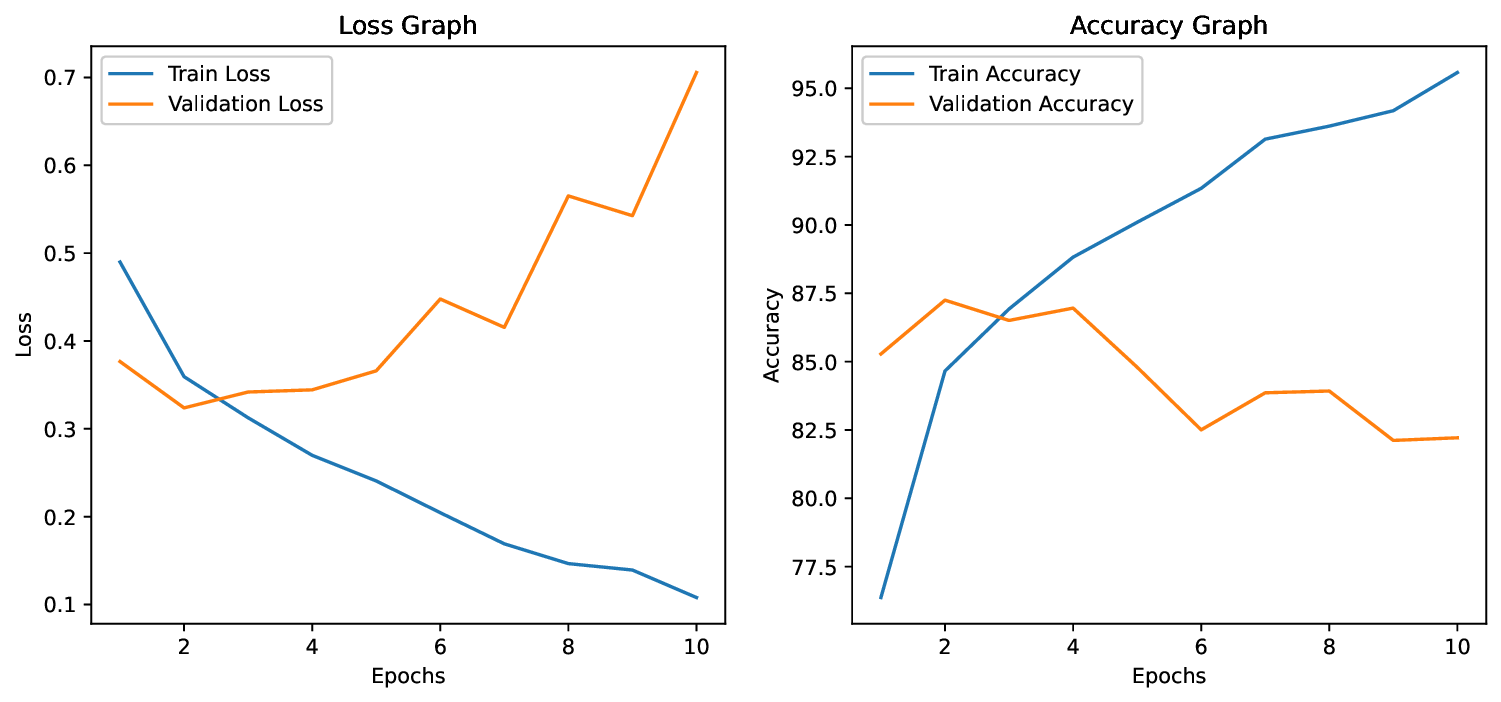}}
	\subfloat[\label{fig10h}]{\includegraphics[width=3cm,height=3cm]{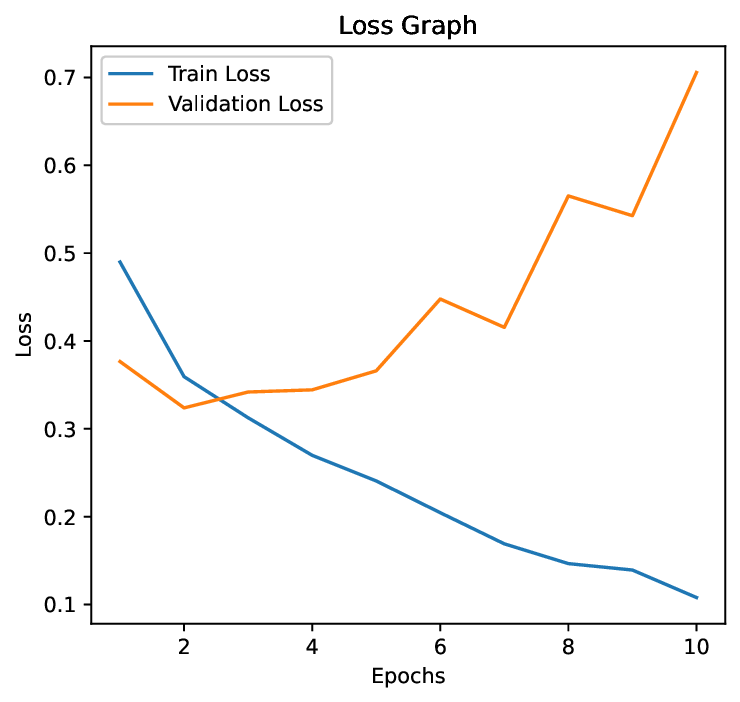}}
	\subfloat[\label{fig10i}]{\includegraphics[width=3cm,height=3cm]{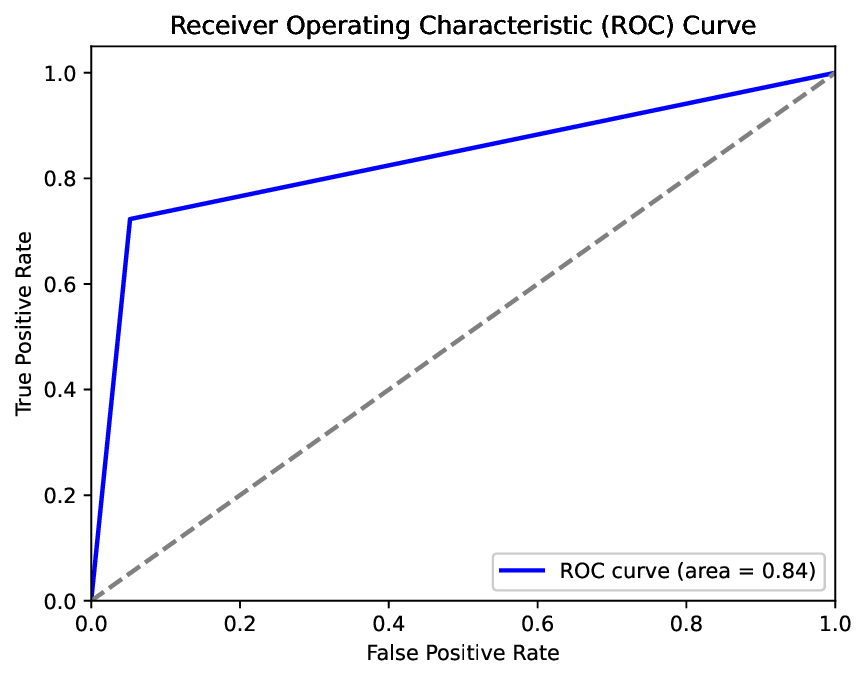}}

	\caption{From top to bottom rows, representing the results of D1+D2 $\rightarrow$ D3, D2+D3 $\rightarrow$ D1 and D3+D1 $\rightarrow$ D2 datasets, respectively. Where D1: NTU-ROSE, D2: ICL and D3: Mturk. From left to right, training and validation accuracy plot, loss plot and testing ROC plot, respectively generated for ten epochs.}
	\label{fig10}
\end{figure}

The training and testing images are from different datasets, which involve different imaging devices and environmental conditions for recapturing. This experiment protocol is the most challenging one. This is the first time cross-domain analysis has been performed on such a large scale. All the previous work focused on intra-domain testing and inter-domain analysis using a single different dataset for testing and training. Combining datasets introduces a more challenging scenario. Also, introducing data augmentation increases the difficulty in classification. The training and validation accuracies and loss can be observed in Fig. \ref{fig10}. The ROCs with the AUC values are also shown in the figure. The table \ref{table8} provides the evaluation details for the cross-domain analysis. The AUC value for the training dataset D1+D2, tested on D3, is the lowest. The other two combinations provide an accuracy of 81\% with precision values of 90\% and 95\%, respectively.

\begin{table}[!htbp]
	\centering
	\caption{Experimental results for the cross domain datasets D1, D2 and D3. $\% \rightarrow \#$  notation denote the training dataset ($\%$) and testing dataset ($\#$). All the evaluation parameters are in \% except the samples which are the count value of images in the training and testing phase.}
	\label{table8}
	\resizebox{\columnwidth}{!}{
		\begin{tabular}{|l|l|l|l|l|l|l|}
			\hline
			\textbf{Datasets} & \textbf{Training Samples} & \textbf{Testing Samples} & \textbf{Accuracy} & \textbf{Precision} & \textbf{Recall} & \textbf{F1-score} \\ \hline
			\textbf{D1+D2 $\rightarrow$ D3} & 5117 & 11432 & 57.24 & 79.87 & 18.35 & 29.85 \\ \hline
			\textbf{D2+D3 $\rightarrow$ D1} &  10644    &  4057    &   81.90    &  90.61     &  81.22   &  85.66     \\ \hline
			\textbf{D3+D1 $\rightarrow$ D2} &  12018    & 2340     &    80.94   &   95.68    &  72.29   &  82.35     \\ \hline
			
	\end{tabular}}
\end{table}

\subsubsection{Comparative Analysis}

Based on the diversity present in the datasets in section \ref{dataset}, we evaluate the performance of different
approaches under the intra, inter and cross-domain experiments in this section. The techniques used for comparative analysis include traditional machine learning classifier with handcrafted features, namely, texture features (LBP$_{8,1}$+SVM \cite{cao2010identification}), aliasing noise and blurring effects (Pixel-wise correlation coeff. + SVM \cite{wang2017simple}), generic CNN models (VGG16 \cite{simonyan2014very}, ResNet 50 \cite{he2016deep} and DenseNet 121 \cite{huang2017densely}). For all the techniques the hyperparameters were empirically selected by the datasets. The feature dimensions for the LBP features was 59 and for the correlation coefficients was 54.

For the intra-database experiment, the results show that the AUCs of most CNN-based models are 0.7. The VGG16 approach achieves AUCs in databases I, II and III, and for inter-domain 64\%. The cross-domain AUCs is between 30-40\%. ResNet approach achieves AUCs for intra-domain and inter-domain  , respectively. The DenseNet achieves AUCs for the three datasets: I, II, and III and for cross-domains . For the LBP-based classifiers, the AUC is 59\% in Database I, 53\% in Database II, and Database III. Given the average performance of all the approaches under an intra-database setting, our method provides the best intra, inter-domain and cross-domain testing results. 

\begin{table*}[!htbp]
	\centering
	\caption{Evaluation on intra, inter and cross-domain datasets. The best performance (accuracy \%) for each framework is bold-faced}
\label{table9}
\begin{tabular}{|l|c|c|c|c|c|c|c|}
	\hline
\textbf{Methods} & \textbf{D1} & \textbf{D2} & \textbf{D3} & \textbf{D1+D2+D3} &\textbf{D1+D2 $\rightarrow$ D3}& \textbf{D2+D3 $\rightarrow$ D1}&
\textbf{D3+D1 $\rightarrow$ D2}\\
\hline
LBP+SVM(RBF) \cite{cao2010identification}&98.95&97.44&93.54&77.40&52.08&57.12&50.50\\ \hline
Pixel-wise Corr. Coef.\cite{wang2017simple} &99.18&97.70&86.75&79.45&50.00&63.21&44.83\\ \hline
VGG16 \cite{simonyan2014very}&79.47&77.60&75.85&64.73&56.63&74.50&76.45\\ \hline
ResNet 50 \cite{he2016deep}&82.50&84.17&82.78&55.78&56.47&75.79&78.03\\ \hline
DenseNet 121 \cite{huang2017densely}&79.12&80.76&81.32&49.17&56.05&55.12&67.95\\ \hline
Proposed DAST &\textbf{99.77}&\textbf{99.65}&\textbf{94.53}&\textbf{81.84}&\textbf{58.24}&\textbf{81.90}&\textbf{80.94}\\
\hline
\end{tabular}
\end{table*}

In this experiment's first part, we use the same image patches for training and testing experimentation. It can be seen from Table \ref{table9} that when the training data and the testing data consist of different domains, the performances of all the detection approaches have decreased significantly compared with those from the intra and inter-database experiments. The AUCs for the LBP-SVM classifier with default parameters are lower than 55\%, which is unsatisfactory. Limitations of LBP features can be overcome by employing other variants of LBP. However, it will increase the feature dimensions. Also, the correlation coefficients for the residual noise images provide accuracy as low as 40\%  for the intra-domain and 63\% for the cross-domain. The traditional feature extraction algorithm is ineffective because of its inability to extract features from different domains. The resolution of the images is different. Also, generalizing the features from tens and hundreds of cameras is impossible using traditional handcrafted features and machine learning classifiers. The problem of underfitting and overfitting is not resolved even after using the regularization parameter.

For the second part, we have employed a few generic CNN models, namely VGG16, ResNet 50, and DenseNet 121; the AUCs on average for the inter and intra-domain are between 75-85\%. The cross-domain results are unsatisfactory. The accuracy is in between 55-75\%. The results achieved by our proposed method are much better. However, all the approaches do not perform well when trained on NTU+ICL image datasets and tested on the Mturk dataset for the following reasons. The testing images are low resolution and less discriminative than the training datasets. The training dataset has a negligible aliasing effect than the testing datasets. The variance in the acquiring device cameras is very high.

\section{Conclusion}
\label{conclusion}
In this work, we proposed a domain generalized recaptured image detection technique, which is a simple but effective anti-forensic scheme using a SWIN transformer. Experimental results have demonstrated that the proposed scheme has good generalization performance under various capturing devices (low and high-resolution cameras), imitating medium (different LCD screens), and different post-processing forgeries (anti-aliasing filters, colour tampering, copy-move and copy-paste). Based on understanding the feature variations, the SWIN transformer extracted the local and global features. The proposed scheme is simple but competitive. The results show that under the most challenging experiment protocol, i.e., the cross-domain experiments, the AUCs of the approaches using the state-of-the-art method are only 0.40 on average, which is way below the results achieved by our model. 
Our study of RIA is limited to images acquired through the ubiquitous LCD screens as the display media. Other rebroadcast imitating mediums,  such as printers, scanners, low-end mobile cameras, and high-quality projection, shall be investigated in the future. Besides detecting the recaptured images, it is also interesting to identify the other recapturing pipelines and estimate the effects of pipeline parameters. Achieving domain generalization for different sources of imitating mediums will be highly challenging.

\bibliographystyle{splncs04}
\bibliography{manuscript_vision}

\end{document}